\documentclass[lettersize,journal]{IEEEtran}
\usepackage{amsmath,amsfonts}
\usepackage{algorithmic}
\usepackage{array}
\usepackage[caption=false,font=normalsize,labelfont=sf,textfont=sf]{subfig}
\usepackage{textcomp}
\usepackage{stfloats}
\usepackage{url}
\usepackage{verbatim}
\usepackage{graphicx}
\usepackage{multirow}
\usepackage{amssymb}
\usepackage{makecell}
\usepackage{threeparttable}
\usepackage{enumerate}
\def\BibTeX{{\rm B\kern-.05em{\sc i\kern-.025em b}\kern-.08em
    T\kern-.1667em\lower.7ex\hbox{E}\kern-.125emX}}
\usepackage{balance}
\begin{document}
\title{SDCM: Simulated Densifying and Compensatory Modeling Fusion for Radar-Vision 3-D Object Detection in Internet of Vehicles}
\author{Shucong Li, Xiaoluo Zhou, Yuqian He, Zhenyu Liu, \textit{Member IEEE}
\thanks{
This work has been submitted to the IEEE for possible publication.
Copyright may be transferred without notice, after which this version may
no longer be accessible.
	
This work was supported in part by the Foshan Social Science and Technology Research Project under Grant 2220001018608, in part by the Guangdong Province Social Development Science and Technology Collaborative Innovation Project under Grant 2023A1111120003, in part by the Guangdong Basic and Applied Basic Research Foundation under Grant 2023A1515012873, and in part by the Guangzhou Key Research and Development Project under Grant 2023B01J0011. (Corresponding author: Zhenyu Liu.)
	
The authors are with the School of Information Engineering, Guangdong University of Technology, Guangzhou 510006, China (e-mail:
lishucong1@mails.gdut.edu.cn; 
2112403196@mail2.gdut.edu.cn;
2112403170@mail2.gdut.edu.cn;
zhenyuliu@gdut.edu.cn; 
)

}}

\markboth{Submitted to IEEE For possible publication,~Vol.~18, No.~9, September~2020}%
{How to Use the IEEEtran \LaTeX \ Templates}

\maketitle

\begin{abstract}
3-D object detection based on 4-D radar-vision is an important part in Internet of Vehicles (IoV). However, there are two challenges which need to be faced. First, the 4-D radar point clouds are sparse, leading to poor 3-D representation.
Second, vision datas exhibit representation degradation under low-light, long distance detection and dense occlusion scenes, which provides unreliable texture information during fusion stage.
To address these issues, a framework named SDCM is proposed, which contains Simulated Densifying and Compensatory Modeling Fusion for radar-vision 3-D object detection in IoV. Firstly, considering point generation based on Gaussian simulation of key points obtained from 3-D Kernel Density Estimation (3-D KDE), and outline generation based on curvature simulation, Simulated Densifying (SimDen) module is designed to generate dense radar point clouds. Secondly, considering that radar data could provide more real time information than vision data,
due to the all-weather property of 4-D radar.
Radar Compensatory  Mapping (RCM) module is designed to reduce the affects of vision datas' representation degradation. Thirdly, considering that feature tensor difference values contain
the effective information of every modality, which could be extracted and modeled for heterogeneity reduction and modalities interaction, Mamba Modeling Interactive Fusion (MMIF) module is designed for reducing heterogeneous and acheiving interactive Fusion. Experiment results on the VoD, TJ4DRadSet and Astyx HiRes 2019 dataset show that SDCM achieves best performance with lower parameter quantity and faster inference speed. Our code will be available.
\end{abstract}


\section{Introduction}
\begin{figure}[t]
	\centering
	\includegraphics[width=3in]{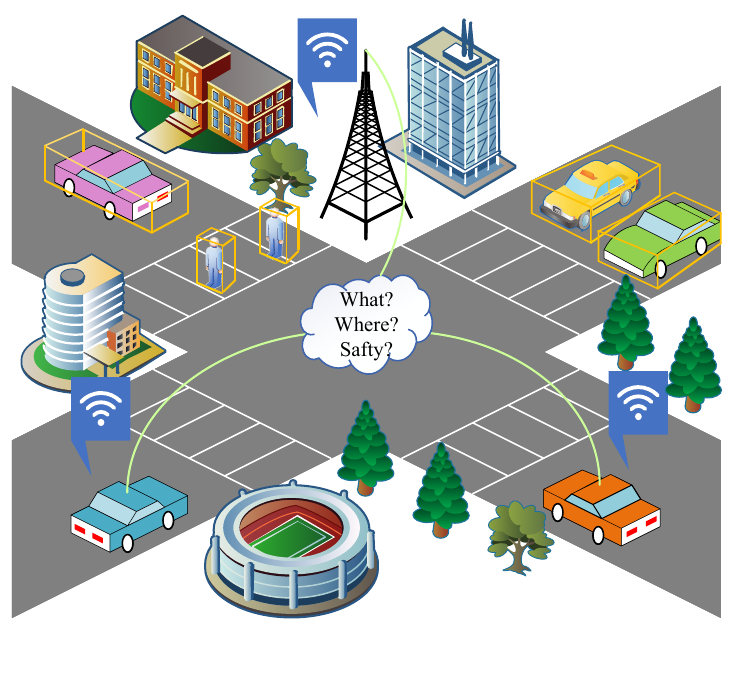}
	\caption{The connection between IoV and 3-D object detection. The communication of correct detection results provides transportation safty information for the vehicles, which is beneficial for making correct driving decisions. \label{iov}
	}
\end{figure}

\IEEEPARstart{I}{nternet} of vehicles has become a necessary part of intelligent transportation system and smart city \cite{smartcity, smartcity1}. In addition, 3-D object detection is one of the sources of traffic safety information in IoV technology \cite{iov}, as shown in Fig \ref{iov}. The vehicles could make correct decisions for real time road condition by the communication of 3-D object detection results. Therefore, the precision of 3-D object detection is important in IoV communication.

To achieve better environment preception, 3-D object detection depends on the collaboration of multiple sensors, including LiDAR, camera and 3-D/4-D millimeter-wave (MMWave) radar \cite{RCBEVDet}. Some sensors show poor performance under adverse weather conditions. The RGB images lacks texture-details under low-light, foggy or night scenes. The LiDAR sensor datas are incompleted easily because LiDAR's near-infrared wave
lengths is limited under rainy or foggy scenes \cite{RCDFNet}. Instead, the 4-D MMWave radar has gradually  become popular for autonomous driving due to the all-weather property, and lower cost than LiDAR. In addition, the 4-D MMWave radar could provide the height information of instances. Therefore, the 3-D object detection adopting 4-D radar has been a hot research with some outstanding achievements recently, which contiains the works based on 4-D radar-vision fusion \cite{LxL, RPFANet} especially.

The 3-D object detection based on 4-D radar-vision fusion could be categoried into concatenation fusion, spatial/channel Attention fusion, cross Attention/Transformer fusion \cite{LxLv2}. The concatenation fusion merges two modalities features through tensor concatenation, and enhances the fusion feature through the variants of Attention \cite{attention} or Transformer \cite{transformer} etc. mechaism, such as LXL \cite{LxL}, Harley et al. \cite{Harley}, HGSFusion \cite{HGSFusion} and so on.
These works ignore the heterogeneity between radar and vision datas, which brings redundant information for the detection. The spatial/channel Attention fusion shows a good balance between computational complexity and precision. RCDFNet \cite{RCDFNet} proposes a variant of channel Attention \cite{CA} for capturing the spatial information and feature representations of modalities' features.
LXLv2 \cite{LxLv2} combines channel and spatial Attention as the second stage fusion for the concatenating features, which is regarded as the first stage fusion. The cross Attention/Transformer fusion is proposed to achieve semantic alignment of modalities' features. RCFusion \cite{RCFusion} designs interactive Attention module based on channel Attention to achieve the complementarity of different modalities. 
ZFusion \cite{ZFusion} proposes deformable cross Transformer module for the dynamic fusion among different modalities, which enhance the robustness of overall framework.

\begin{figure}[t]
	\centering
	\includegraphics[width=3.5in]{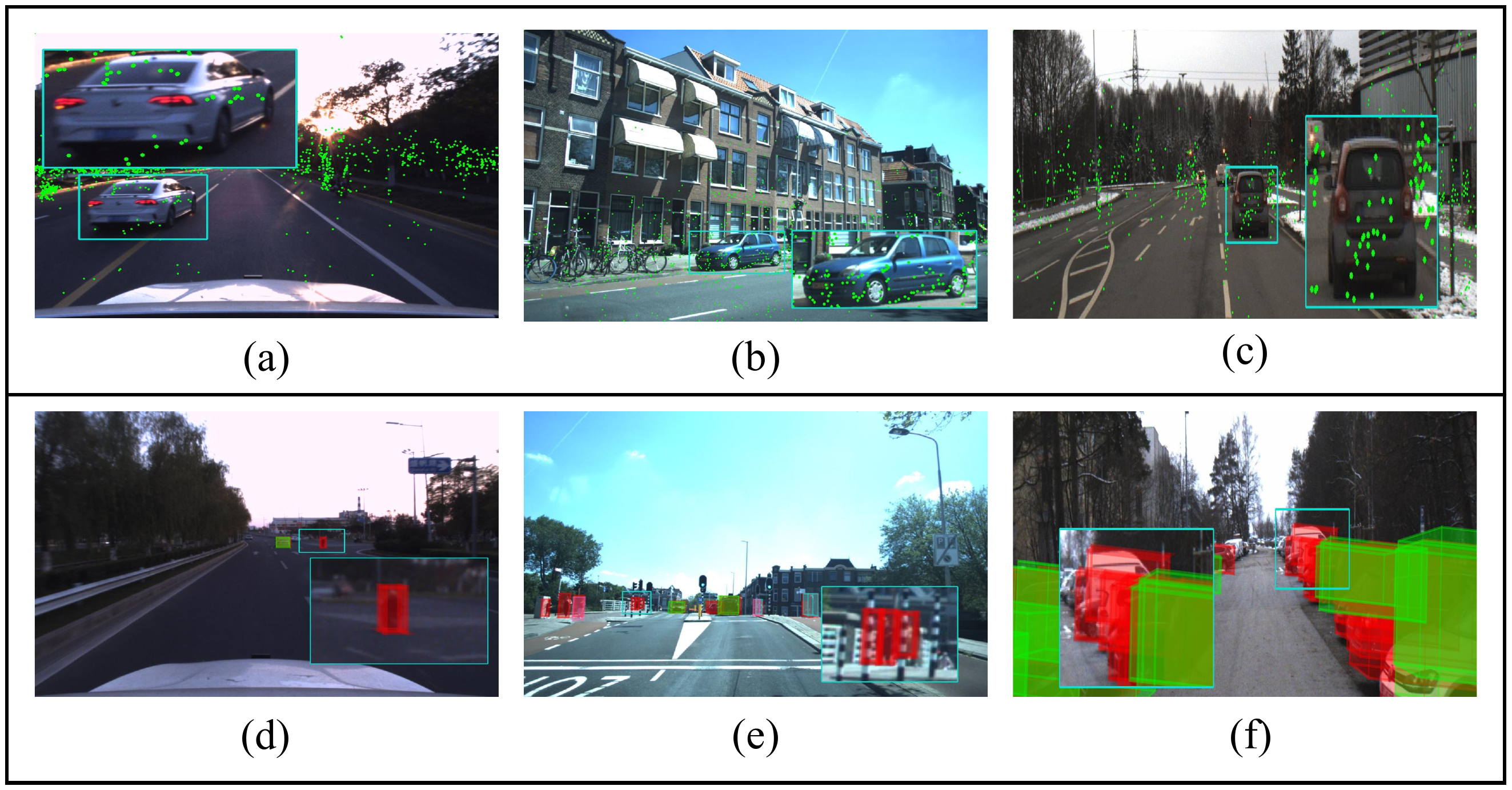}
	\caption{The challeges of 3-D object detection based on 4-D radar-vision fusion. In the first row, (a), (b) and (c) shows the sparsity of 4-D radar point clouds, the instances have few points generated by 4-D radar sensors. The second row shows the missed detection of state-of-the-art methods, which are  nder the representation degradation of vision datas. (d) is low-light, (e) is long distance detection, and (f) is dense occlusion. The red bounding boxes are GroundTruth, while other color bounding boxes are the detection results. These images come from VoD, TJ4DRadSet and Astyx HiRes 2019 dataset.  \label{problem}
	}
\end{figure}

However, although the above works obtain outstanding achievements, there are two challeges being needed to face. First, the 4-D radar point clouds are sparse, leading to poor 3-D representation \cite{jiang, luan}, as shown in Fig \ref{problem} (a), (b) and (c). 
Second, the vision datas have representation degradation under low-light, long distance detection and dense occlusion scenes, which bring unreliable texture information for multi-modality interactive fusion, as shown in Fig \ref{problem} (d), (e) and (f) respectively. The vision datas have limited texture-details under low-light \cite{Wavemamba}. In addition, there are few pixels on the instances at long distance \cite{pixels}. Moreover, the dense occlusion loses more features about the instances \cite{occlusion}. These scenes bring representation degradation of vision datas.

To this end, a 3-D object detection based on 4-D radar-vision fusion framework named SDCM is proposed, which contains three parts:
\begin{enumerate}[(1)]
	\item To face the problem of 4-D radar point clouds whose sparsity leads to poor 3-D representation of objects, SimDen module is designed. The surface of objects are completed by Gaussian simulation based on key points, which are obtained by 3-D KDE and depth filtering. The outline of objects are generated by curvature simulation, which referring the curvature esitmation of edge points.
	\item To reduce the affect of representation degradation from vision datas, RCM module is designed. The radar features are adopted to compensate for the deficiencies caused by the degradation of visual features, after the enhancement under several sizes of receptive field.
	\item To reduce the heterogenity and achieve feature interaction fusion, MMIF module is designed. The effective information in the feature tensor difference values is extracted and modeled by the long-ranges dependencies capture and linear complexity of Mamba mechanism, and the interaction fusion could be achieved by inquiring effective information between the two modalities.

\end{enumerate}
Overall, our contributions could be summarized as follows:
\begin{itemize}
	\item SDCM is proposed, a novel 3-D object detection framework based on 4-D radar-vision fusion, which contains simulated densifying and degradation perception fusion. 
	\item SimDen module is designed to achieve 4-D radar point clouds densifying through Gaussian simulation for surface completing and curvature simulation for outline generation.
	\item RCM module is designed to reduce the affect of representation degradation from vision datas through radar features compensation under several sizes of receptive field.
	\item MMIF module is designed to reduce the heterogenity and achieve feature interaction fusion through obtaining and modeling the effective information of two modalities from feature tensor difference values. 
	\item The experiment results on VoD \cite{VoD}, TJ4DRadSet \cite{TJ4DRadSet} and Astyx HiRes 2019 \cite{Astyx} datasets shows that SDCM achieves best performance with lower parameter quantity and faster inference speed.
\end{itemize} 

\section{Related Work}
\subsection{4-D Radar Point Clouds Densifying}
To slove the problem of 4-D radar point clouds' sparsity, many works devote themselves to conduct point clouds densifying before the feature learning stage of detectors. SMURF \cite{SMURF} employs a spatial multi-representation fusion approach that includes pillarization and KDE with settled bandwidths to densify the radar point clouds. Wang et al. \cite{Wang} adopts Nearest Neighbor Interpolation on the segmetation masks of the instances, which contains several radar points. HGSFusion \cite{HGSFusion} introduces hybrid probability density distribution to densify the radar point clouds, as well as decrease the error from Direction of Arrival algorithm. RADEN \cite{RADEN} adopts geometrical domain densifying to increase the number of global radar point clouds, the geometrical features of local instances are recovered.  

However, although these works generate dense radar point clouds, they ignore the geometrical information of the instances, like Wang et al. and HGSFusion, or they introduce noise from global perspective, like SMURF and RADEN.

\subsection{3-D Object Detection Based On  Radar Or Vision}
The vision-based 3-D object detection methods focus on BEV maps in recent years, with the emergence of multiview technelogy.
BEVDepth \cite{BEVDepth} employs LiDAR to obtain ground-truth depth, predicts a dense depth distribution at the pixel level, and then converts it into the BEV representation. BEVNeXt \cite{BEVNeXt} utilizes CRF-modulated depth estimation from multi-view images, generates a dense BEV representation, aggregates temporal information via Res2Fusion.
Other works like Wang et al.\cite{DETR3D}, BEVFormer \cite{BEVFormer} and DualBEV \cite{DualBEV} also achieve good performances on vision-based 3-D object detection.

However, these works contains the emergence of spatial distortion, because the RGB images lack depth information of objects. In addition, the vision datas lacks enough texture-details under the low-light enviroments.

The radar-based 3-D object detection become a hot reserach gradually due to the all-weather property of radar sensor, especially for 4-D MMWave radar, whose point clouds are denser than 3-D MMWave radar, and the doppler information are provided. SMURF \cite{SMURF} enhances radar point cloud representations through pillarization and KDE to counteract accuracy degradation from multipath effects. MUFASA \cite{MUFASA} enhances radar point cloud feature extraction by leveraging the GeoSPA and DEMVA modules to capture local and global information. MAFF-Net \cite{MAFF-Net} leverages sparsity pillar Attention and  cluster query cross-Attention to enrich object features and suppress noise, employing cylindrical denoising assist to refine bounding box predictions.

However, these works are limited in object classification due to the absence of texture-details, which could not be provided by MMWave radar.

\subsection{3-D Object Detection Based On Radar-Vision Fusion }
The high complementarity of radar and vision datas brings Attention for the research of radar-vision 3-D object detection. The radar datas could provide the depth, location and doppler information, while the vision datas could provide texture-details about the objects.
RCFusion \cite{RCFusion} proposes cross-Attention mechanism for radar-vision feature fusion initially. UniBEVFusion \cite{UniBEVFusion} enhances 3-D detection by integrating radar depth with a unified Lift-Splat-Shoot \cite{LSS} (LSS) module and fusing multi-modal features through shared encoders. RCDFNet \cite{RCDFNet} integrates radar and camera data through dual-level fusion and converts them into unified BEV representations for robust 3-D detection. SGDet3D \cite{SGDet3D} proposes  dual-branch fusion module and object-oriented Attention module to convert the fused features into a unified BEV representation for 3-D object detection. 

However, these works ingore the representation degradation of vision datas under low-light, long distance detection and dense occlusion scenes. The degradation brings unreliable texture information for radar-vision interactive fusion. In addition,
these works have high computational complexity, which brings the constraint of real time deployment.

\begin{figure*}[t]
	\centering
	\includegraphics[width=7in]{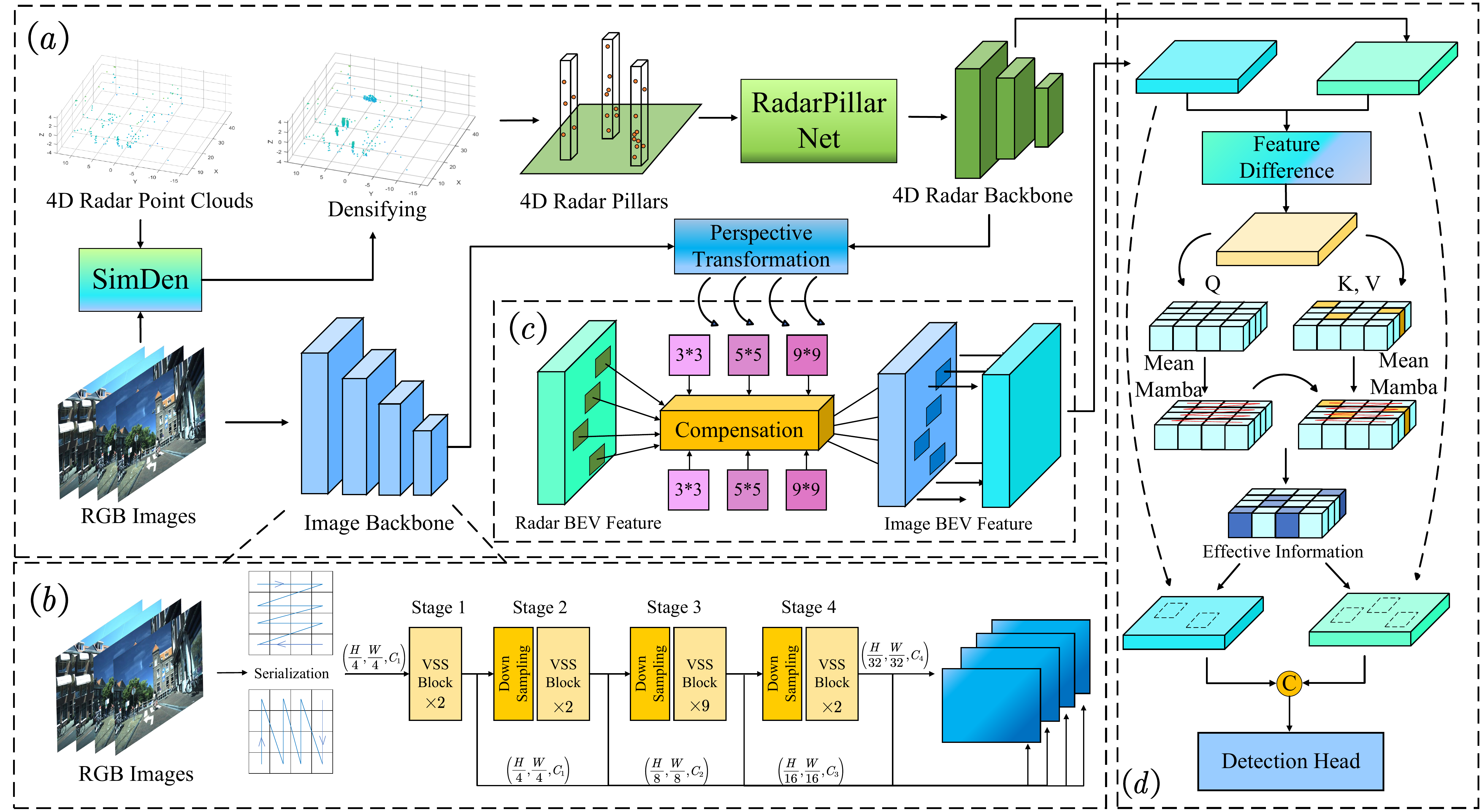}
	\caption{The overall pipeline of SDCM. (a) contains SimDen module, as well as the backbones of radar point clouds and RGB images. (b) is VMamba architecture which is adopted as image backbone. (c) is RCM module, which adopts radar features to compensate the represenstation degradation of vision datas. (d) is MMIF module, which reduce heterogeneous features and achieve feature interaction fusion. \label{pipeline}
	}
\end{figure*}
\begin{figure*}[t]
	\centering
	\includegraphics[width=7in]{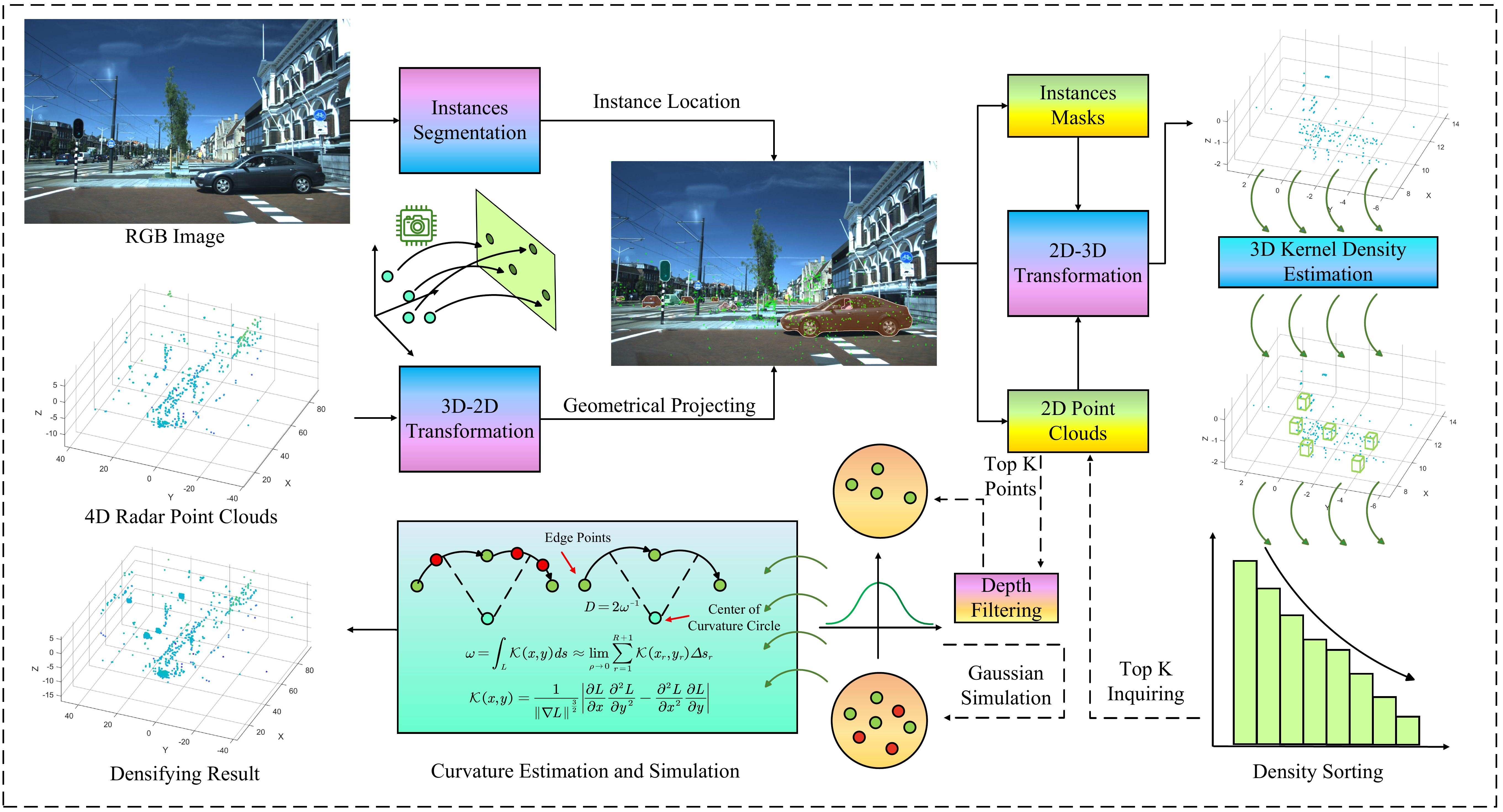}
	\caption{The overall architecture of SimDen module. The dense radar point clouds are simulated through two steps, one is 3-D KDE and Gaussian simulation, the other is curvature estiamtion and simulation.  \label{simden}
	}
\end{figure*}
\section{Methodology}
The overall pipeline of SDCM is shown in Fig \ref{pipeline}. The RGB images and original sparse radar point clouds are conveyed to SimDen module for point clouds densifying, then the dense point clouds and RGB images are conveyed to their backbones to achieve feature learning. After that, the two features are sent to RCM module for the  compensation of representation degradation, as shown in Fig \ref{pipeline} (c). Finally, the two features are conveyed to MMIF module for reducing heterogeneous features and achieving feature interaction fusion, as shown in Fig \ref{pipeline} (d).
\subsection{Simulated Densifying Module}
The radar point clouds on the objects represent real measurment information such as geometrical location, doppler value and so on. Therefore, these points could be regarded as prior knowledge to generate new point clouds.
The key points could be found out through referring their density values within these points, which are passed through depth filtering for noise removing \cite{MVP, HGSFusion}. The new point clouds could be generated based on key points, following a Gaussian distribution pattern so that the surfaces of objects could be completed. In addition, 
it is necessary to ensure that the object point clouds has smooth outlines, which could be generated by referring curvatures between the points locating on the edges.
The positions of outline points are determined by the magnitude of curvature values. 

Based on above ideas, SimDen module is proposed to generate dense radar point clouds for 3-D object detection, as shown in Fig \ref{simden}. The radar point clouds $ P_{3D}\in \mathbb{R}^{N\times 3}$ are projected on the RGB images $img\in \mathbb{R}^{H\times W}$ through intrinsic and extrinsic parameters matrix, denoted as $D_I$ and $D_E$ respectively. To obtain the points of instances, the RGB images are fed to segmentation model for seperating the instances from background. Therefore, the 2-D point clouds within the segmentation masks, denote as $ P_{2D}^{Ins}$, could be regarded as prior knowledge of real measurment information tentatively, this program could be described as:
\begin{equation}\label{equ1}
	\begin{split}
	P_{2D} &=D_ID_EP_{3D}\\
	M&=Seg\left( img \right) \\
	\mathcal{Q}&=\bigcup_{n=1}^N{\left( M_n\cap P_{2D} \right)}\\
    \end{split}
\end{equation}
where $ Seg(\cdot)$ denotes segmentation model's inference function, and $M$ denotes the masks of $N$ objects. The points are filtered to remove the noise and non-object poinbts by the mode of depth values:
\begin{equation}\label{equ1c}
	\begin{split}
&P_{2D}^{Ins}=\\
&\left\{ \left( x_n,y_n,d_n \right) |\lfloor d_n \rfloor =\lfloor mode\left( \mathcal{Q}_n\left[ :,2 \right] \right) \rfloor ,\ \forall n\in \left[ 1,\ N \right] \cap \mathbb{Z} \right\}
    \end{split}
\end{equation}
where $x$, $y$ denote the coordinate values of $ P_{2D}^{Ins}$.
$\lfloor \cdot \rfloor$ denotes downward rounding function and $mode(\cdot)$ denotes mode value function, which is adopted for depth values $d$ filtering to obtain the real 2-D points on the instances.

The $ P_{2D}^{Ins}$ should be transformed into 3-D Cartesian space for density caculation, whose aim is finding out the key points. The density calculation relies on the point group within a certain range of the key points, which could be achieved by KDE. The Gaussian kernel function is adopted for the density calculation of each points in $ P_{2D}^{Ins}$, the density value could be described as:
\begin{equation}\label{equ2}
	\begin{split}
&K\left( p_i \right)=\\ &\frac{1}{W\displaystyle\prod_{j=1}^J{B_j}}\sum_{w=1}^W{\exp \left\{ \frac{-\gamma}{2}\left[ p_i-E\left( p_i \right) \right] _{w}^{T}\left[ p_i-E\left( p_i \right) \right] _w \right\}}
\end{split}
\end{equation}
where $W$ denotes the number of 3-D points on the objects, $J$ denotes the number of 3-D points' dimensions. $\gamma$ denotes the scaling factor with default value 1.0. $ E(\cdot)$ denotes the mathematical expectation function. $p_i$ denotes the $i$-th 3-D point of the $n$-th objects, which are transformed into 3-D Cartesian space from  $ P_{2D}^{Ins}$, denotes as $ P_{3D}^{Ins}$, this program could be described as:
\begin{equation}\label{equ3}
\begin{split}
	&P_{3D}^{Ins}=D_{E}^{-1}D_{I}^{-1}P_{2D}^{Ins}\\
	&p=\left\{ p_i|p_i\in \left[ P_{3D}^{Ins} \right] _n,\ \forall i\in \left[ 1,\,\,I \right] \cap \mathbb{Z}, \ \forall n\in \left[ 1,\,\,N \right] \cap \mathbb{Z} \right\} 
\end{split}
\end{equation}
 $B_j$ denotes the bandwith of the $j$-th dimentions of 3-D radar point clouds. The bandwidths of all dimensions are consistent, and the values are calculated through experience rules \cite{silverman,scott} or user-defined constants $\mathcal{B}$:
\begin{equation}\label{equ4}
	B_j=\begin{cases}
		W^{\frac{-1}{J+4}},&		\text{Scott\,\,Rule}\\
		\left[ \frac{W\left( J+2 \right)}{2} \right] ^{\frac{-1}{J+4}},&		\text{Silverman\,\,Rule}\\
		\mathcal{B},&		\text{User-Defined}\\
	\end{cases}\ \forall j\in \left[ 1,\,\,J \right] \cap \mathbb{Z}	
\end{equation}

 For all of 3-D points, their inputs of estimation is Manhattan distance matrices, whose type is zero-diagonal symmetric matrix. The input is denoted as $H$ and described as:
 \begin{equation}\label{equ5}
 H\left( u,v \right) =\begin{cases}
 	0,&		u=v\\
 	d_{uv}=d_{vu},&		u\ne v\\
 \end{cases}
\end{equation}
where the elements of $H$ is calculated by:
\begin{equation}\label{equ6}
d_{uv}=d_{vu}=\sum_{j=1}^J{\left| u_j-v_j \right|},\ \forall u,v\in \left[ 0,\,\,W \right) \cap \mathbb{Z}
\end{equation}
Therefore, the key points are chosen by sorting the magnitude of density values.
 However, due to the extremely sparse characteristic, some objects may contain only 1 or 2 points as their prior knowledge. For these situations, the maximum number of key points is set to $\mathcal{T}$. The key points' center is regarded as the source of densifying. The centers could be obtained through:
\begin{equation}\label{equ7}
	p_{c}^{3D}=\frac{1}{T}\left( \sum_{t=1}^T{x_t},\ \sum_{t=1}^T{y_t},\ \sum_{t=1}^T{z_t} \right) ,\ \forall T\in \left[ 1,\mathcal{T} \right] \cap \mathbb{Z} 
\end{equation}
where $x$, $y$ and $z$ denote the 3-D Cartesian coordinate values of key points, $T$ denotes the number of key points.
The new points $P_{new}$ could be generated around the center by the Gaussian simulation, their probability density function is described as:
\begin{equation}\label{equ8}
\begin{split}
	&f\left( P_{new}|\mu ,\ \varSigma \right) =\\ 
	&\frac{1}{\left( 2\pi \right) \left| \varSigma \right|^{\frac{1}{2}}}\exp \left[ -\frac{1}{2}\left( P_{new}-\mu \right) ^T\varSigma ^{-1}\left( P_{new}-\mu \right) \right]
\end{split}
\end{equation}
where $\mu$ denotes the mean of Gaussian distribution and $\varSigma$ denotes the covariance matrix, they satisfy:
\begin{equation}\label{equ9}
	\begin{split}
      &\mu =p_{c}^{2D}=\frac{D_ID_E}{T}\left( \sum_{t=1}^T{x_t},\,\,\sum_{t=1}^T{y_t},\,\,\sum_{t=1}^T{z_t} \right)  \\
      &\varSigma =E\left[ \left( P_{new}-\mu \right) \left( P_{new}-\mu \right) ^T \right] 
\end{split}
\end{equation}

At this step, the new point clouds on the objects do not have obvious outlines. To reduce outliers and obtain smooth outlines, the points locating on the edge of objects are adopted for curvature estimation and simulation. $R$ referring points are inserted between any two edge points.
By this way, the curvature $\omega$ of these two points could be obtained by:
\begin{equation}\label{equ10}
	\begin{split}
		&\omega =\int_L{\mathcal{K}\left( x,y \right) ds}\approx \underset{\rho \rightarrow 0}{\lim}\sum_{r=1}^{R+1}{\mathcal{K}\left( x_r,y_r \right)}\varDelta s_r\\
		&\rho =\max \left( \varDelta s_1,\varDelta s_2,\cdots ,\varDelta s_r,\cdots ,\varDelta s_R \right) \\
		&\mathcal{K}\left( x,y \right) =\frac{1}{\lVert \nabla L \rVert ^{\frac{3}{2}}}\left| \frac{\partial L}{\partial x}\frac{\partial ^2L}{\partial y^2}-\frac{\partial ^2L}{\partial x^2}\frac{\partial L}{\partial y} \right|
	\end{split}
\end{equation}
where $L$ is the undirected curve passing through all the referring points between two points. $\varDelta s_r$ denotes the $r$-th unit curve of $L$. Actually, the line integrals of the first type is calculated through the form of Riemann sum. The curvature between three points could be calculated using the radius of their circumcircle, which could be obtained by Sine Theorem:
\begin{equation}\label{equ11}
	\frac{1}{\omega _r}=\frac{2\varDelta s_r}{2\sin \theta _r}=\frac{\varDelta s_r}{\sqrt{1-\cos ^2\theta _r}}
\end{equation}
where $\omega _r$ and $\theta _r$ denote the curvature and angle of the curve centered at the $r$-th point respectively.
The cosine value of $\theta _r$ could be calculated following Cosine Theorem. Therefore, the whole curvature could be obtained under the condition of integral interval's equal division:
\begin{equation}\label{equ12}
	\begin{split}
		&\forall r\in \left[ 1,R \right] \cap \mathbb{Z},\ \exists \rho =\varDelta s_r,\ s.t.\\
		&\omega \approx \underset{\rho \rightarrow 0}{\lim}\sum_{r=1}^{R+1}{\frac{\sqrt{1-\cos ^2\theta _r}}{\varDelta s_r}}\varDelta s_r=\sum_{r=1}^{R+1}{\sqrt{1-\cos ^2\theta _r}}
	\end{split}
\end{equation}
The outline points $P_o$ could be simulated by equidistant interpolation, which is based on curvature. If the curvature is too big, the outline points should be near to the edge points. Conversely, its location could be near to the center. The simulation could be described as:
\begin{equation}\label{equ13}
P_o=\frac{p_r+\omega p_{r-1}}{1+\omega},\ \forall r\in \left[ 1,\ R \right] \cap \mathbb{Z}
\end{equation}	
where $p_r$ denotes the $r$-th referring points. Therefore, the simulated 2-D point clouds are conbined with $P_{new}$ and $P_o$, which should be transformed into 3-D Cartesian space as simluated 3-D point clouds $P_{sim}^{3D}$:
\begin{equation}\label{equ14}
P_{sim}^{3D}=D_{E}^{-1}D_{I}^{-1}\left( P_{new}\cup \,P_o \right)  
\end{equation}

\begin{figure*}[t]
	\centering
	\includegraphics[width=7in]{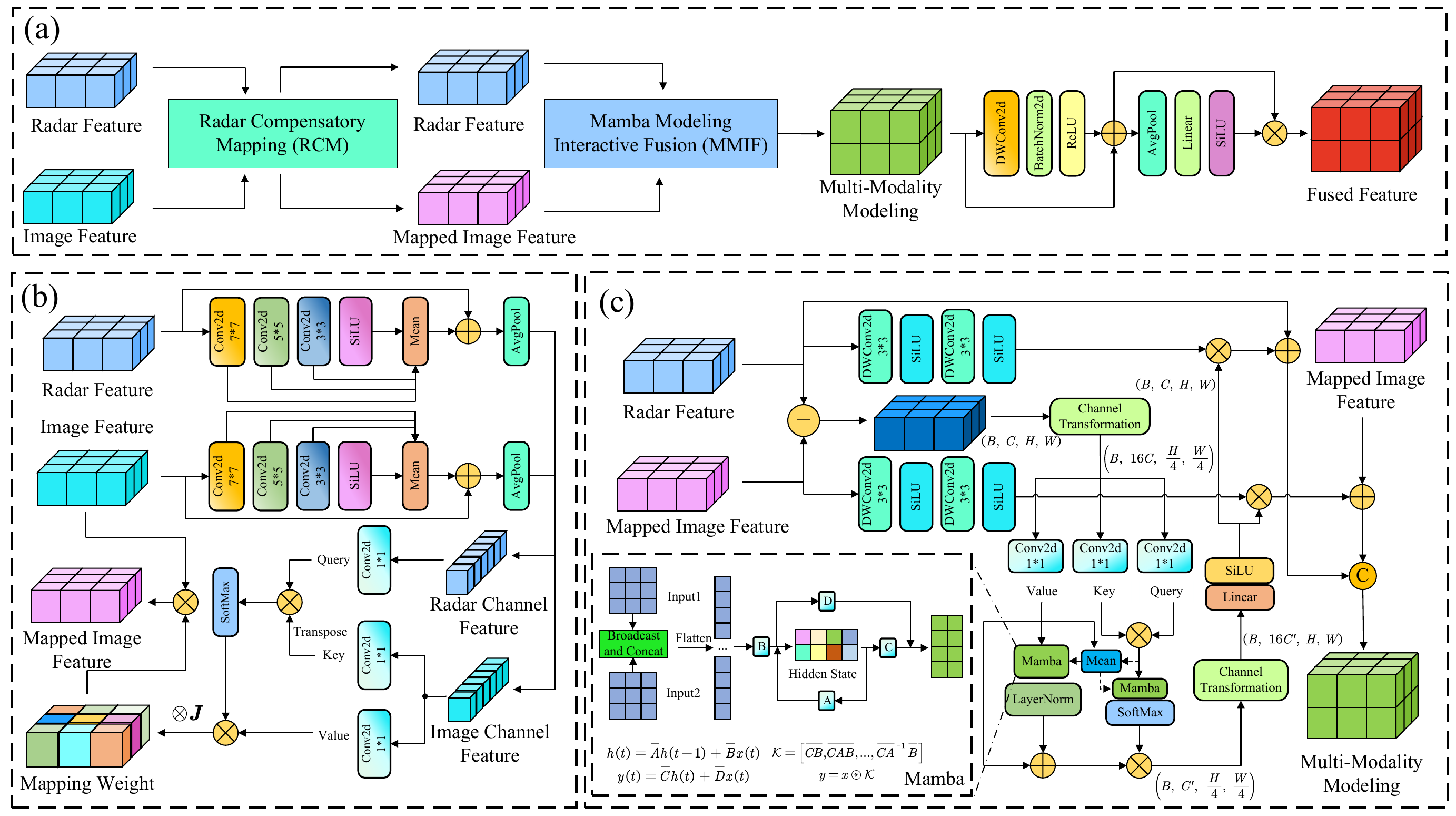}
	\caption{The overall architecture of fusion stage in SDCM. (a) is the program of the fusion stage. (b) is the RCM module, which achieves the compensation of representation degradation. (c) is the MMIF module, which reduces heterogeneous features and achieves feature interaction fusion. \label{usmdmf}
	}
\end{figure*}
\subsection{Radar Compensatory Mapping Module } 
The radar data could provide more real time information than RGB images, such as geometrical location, doppler value and so on. Therefore, under the scenes of low-light, long distance detection and dense occlusion, the information of objects in two modalities' features should be enhanced under several sizes of receptive field, then the
radar features could be adopted to compensate for the deficiencies caused by the degradation of visual features through feature compensation.

Based on the above idea, RCM module is designed for the compensation of representation degradation from vision datas, as shown in Fig \ref{usmdmf} (b). The features of two modalities come from their backbones, they are denoted as $F_{radar}$ and $F_{img}$ respectively. The two features are conveyed to several convolution layers with a group of receptive fields, the purpose of this step is to capture different sizes of objects in the real time road scenes. This program could be described as:
\begin{equation}\label{equ15}
\overline{F}_m=\frac{1}{Z}SiLU\left[ \sum_{z=1}^Z{Conv2D}\left( F_{z-1},\ ker_z \right) \right] +F_m
\end{equation}
where $F_m$ is $F_{radar}$ or $F_{img} $, $Z$ denotes the number of receptive fields, and $ker$ denotes the size of convolution kernel. $Conv2D(\cdot)$ and $SiLU(\cdot)$ denote 2-D convolution and SiLU activation function respectively.

The deep level features are embedded within the channel information. Therefore, the two features are conveyed to Average Pooling layers to obtain channel features. The image channel feature should inquire the objects' information from radar features. Therefore, the radar channel feature is regarded as query to inquire the information of image channel feature, which is regarded as key and value. This program is described as:
\begin{equation}\label{equ16}
	\begin{split}
		Q\ &=\ Conv2D\left[ AP\left( \overline{F}_{radar} \right) ,\,\,ker=1 \right] \\
		K\ &=\,Conv2D\left[ AP\left( \overline{F}_{img} \right) ,\,\,ker=1 \right] \\
		V\ &=\,\,Conv2D\left[ AP\left( \overline{F}_{img} \right) ,\,\,ker=1 \right]
	\end{split}	 
\end{equation}
where $AP(\cdot)$ denotes Average Pooling function.
Following the Attention mechanism, the image feature inquires the objects' information from radar data by multiplying with mapping weight. This program could be described as:
\begin{equation}\label{equ17}
    F_{Mimg}=F_{img}\odot \left[ SoftMax\left( \frac{QK^T}{\sqrt{d_k}} \right) V\otimes \boldsymbol{J}_{H_1\times W_1} \right] 
\end{equation}
where $SoftMax(\cdot)$ denotes softmax activation function. $d_k$ denotes the dimensions of $K$, $\boldsymbol{J}_{H_1\times W_1}$ denotes a $H_1\times W_1$ matrix whose all elements are 1. $\odot$ and $\otimes$ denote Hadamard and  Kronecker product operation respectively.

\subsection{Mamba Modeling Interactive Fusion Module}
The modalities' features outputed from RCM module remain heterogenity because it only achieves the compensation of objects' information. To this end, the feature tensor difference values could be adopted for heterogenity reduction, because they reflect the effective information which could be provided by every modality individually \cite{valid, valid1}.
Therefore, the heterogenity could be reduced if modalities provide each other with effective information.

The effective information could be concentrated on the channel dimension and extracted by Mamba \cite{Mamba} mechanism, based on whose long-ranges dependencies capture and linear complexity. The object of Mamba's scanning is the entirety of feature maps and their mean tensors which contain channel information, because the deep modeling could be achieved under the assistance of channel information. By this way, the weights of effective information are generated, which could be provided by modality's feature to another modality for enhancement interactively. Therefore, the 
heterogenity of two modalities is reduced.

Based on the above ideas, MMIF module is designed for reducing heterogeneity and acheiving features interaction fusion. The two features are calculated the feature tensor difference values $\varDelta F$ and then conveyed to several DWConv2D-SiLU layers for features enhancement. This program could be described as:
\begin{equation}\label{equ18}
	\begin{split}
		\varDelta F&=F_{radar}-F_{Mimg}\\
		\widetilde{F}_{m}&=f_G\in \left\{ f_g|f_{g+1}=SiLU\left[ DWConv2D\left( f_g, ker=3\right) \right] \right\} \\ 
	\end{split}
\end{equation}
where $\widetilde{F}_{m}$ denote the enhanced features of radar and image data, which could be denoted as $\widetilde{F}_{radar}$ and $\widetilde{F}_{Mimg}$. $G$ is the number of DWConv2D-SiLU layers and satisfies $\forall g\in \left[ 0,\ G \right] \cap \mathbb{Z}$. $SiLU\left[ DWConv2D\left( \cdot \right) \right]$ denotes the DWConv2D-SiLU layer.

The effective information in the residual features is concentrated on the channel dimension, which is achieved by channel reshaping. The reduction in the size of feature map leads to a decrease in its resolution, which is hard for tradictional Attention mechanism to obtain interactive weights.
Therefore, the Mean-Mamba block is designed for  
extracting deep effective information from limited resolution of feature maps. This design depends on the long-ranges dependencies capture and linear complexity of Mamba.
The mean of feature maps are adopted as a part of input for the scanning of Mamba because deep modeling could be achieved under the assistance of channel information, which could be described as:
\begin{equation}\label{equ19}
\overline{I}=\left[ \frac{1}{H_2W_2}\sum_{h=1}^{H_2}{\sum_{w=1}^{W_2}{I\left( :,\,\,h,\,\,w \right)}} \right] \otimes \boldsymbol{J}_{H_2\times W_2}
\end{equation}
where $I$ and $\overline{I}$ denote the input data and its channel information. $H_2$ and $W_2$ denote the size of feature maps after channel transformation.

The scanning of Mamba focuses on the entirety of $I$ and $\overline{I}$, which is serialized through Z-order along the longitudinal and transverse directions. Following the ZOH Principle  \cite{Hippo} adopted in Mamba machanism, the learnable parameters $A$, $B$, $C$ and $D$ under the continuous state are transformed into discrete state, and take part in the iteration of hidden state $\overline{h}$.
This program could be described as:
\begin{equation}\label{equ20}
	\begin{split}
	\overline{s}_t\in \overline{S}&=ZO\left( I \right) \cup ZO\left( \overline{I} \right)\\
	\overline{h}\left( t \right) &=\overline{A}\overline{h}\left( t-1 \right) +\overline{B}\overline{s}_t\\
	Y\left( t \right) &=\overline{C}\overline{h}\left( t \right) +\overline{D}\overline{s}_t\\
	\end{split}
\end{equation}
where $ZO(\cdot)$ denotes the Z-oder serialization operator. $t$ denotes the time step and $\overline{s}_t$ denotes the tokens in the entirety of $I$ and $\overline{I}$. $Y$ denotes the output of Mamba.

The input datas are the linear mapping of residual features. The Value is modeled through the program of:
\begin{equation}\label{equ21}
	\begin{split}
		&\varDelta F_{H_2\times W_2}=CT\left( \varDelta F_{H_1\times W_1} \right) \\
		&\mathcal{V}=LN\left( \overline{S}_v\circledast \mathcal{K}_{Mamba} \right) +\boldsymbol{W}_1\varDelta F_{H_2\times W_2}
	\end{split}
\end{equation}
where $ CT(\cdot)$ denotes Channel Transformation operator, it reduces the size of feature maps and inprove the number of channels. $LN(\cdot)$ denotes the LayerNorm operator, $\overline{S}_v$ denotes the input sequence of Value following Eq \ref{equ20}. $\circledast$ denotes convolution operator.  $\boldsymbol{W}$ denotes the linear mapping weight. $\mathcal{K}_{Mamba}$ denotes the convolution type of Mamba machanism, which is described as:
\begin{equation}\label{equ22}
	\begin{split}
	\mathcal{K}_{Mamba}&=\left[ \overline{C}\overline{B},\ \overline{C}\overline{A}\overline{B},\ \cdots ,\ \overline{C}\overline{A}^{-1}\overline{B} \right] \\
	\overline{A}&=\exp \left( \Delta A \right) \\
	\overline{B}&=\left( \Delta A \right) ^{-1}\left( \overline{A}-\boldsymbol{E} \right) \left( \Delta B \right) 
	\end{split}
\end{equation}
where $\Delta$ denotes the step size of ZOH Principle.

The key and query are fused together for the modeling of Mamba, which could be described as:
\begin{equation}\label{equ23}
	\begin{split}
		I_{kq}&=c^{-\frac{1}{2}}\boldsymbol{W}_2\varDelta F_{H_2\times W_2}\left( \boldsymbol{W}_3\varDelta F_{H_2\times W_2} \right) ^T\\
		\overline{S}_{kq}&=ZO\left( I_{kq} \right) \cup ZO\left( \overline{I}_{kq} \right) \\
		\mathcal{W}&=\overline{S}_{kq}\circledast \mathcal{K}_{Mamba}
	\end{split}
\end{equation}
where $ c$ denotes the number of channels of key values. Therefore, the output $\mathcal{E}_{H_1\times W_1}$ contains deep effective information about two modalities, whose enhanced features could achieve interaction by inquiring the output. To prevent vanishing gradient problem, the original features should be added into fusion through shortcuts. Therefore, the fused feature $\widehat{F}$ could be achieved by concatenating $Cat(\cdot)$. This program could be described as:
\begin{equation}\label{equ24}
	\begin{split}
		\mathcal{E}_{H_1\times W_1}&=CT\left[ SoftMax\left( \mathcal{W} \right) \mathcal{V} \right] \\
		\widehat{F}_{radar}&=\widetilde{F}_{radar}\odot \mathcal{E}_{H_1\times W_1}+F_{radar}\\
		\widehat{F}_{img}&=\widetilde{F}_{Mimg}\odot \left( \boldsymbol{J}_{H_1\times W_1}-\mathcal{E}_{H_1\times W_1} \right) +F_{Mimg}\\
		\widehat{F}&=Cat\left( \widehat{F}_{radar},\ \widehat{F}_{img} \right) 
	\end{split}
\end{equation}

\subsection{Loss Function}
The joint loss function of SDCM contains four parts.

The first part is classification loss, which is calculated by Sigmoid Focal Loss \cite{focal}:
\begin{equation}\label{equ25}
	\mathcal{L}_{cls}=-\alpha _t\left( 1-p_t \right) ^{\sigma}\log _2\left( p_t \right) 
\end{equation}
where $t$ denotes the index of categories, $\alpha_t$ and $p_t$ denote the weight factor and class probability of $t$-th class respectively. The value of $\alpha$ and $\sigma$ are set to 0.25 and 2 during training.

The second part is Radar Occupancy Loss $\mathcal{L}_{occ}$, which is adopted for semantic classfication of voxels or pillars. The calculation is based on Sigmoid Focal Loss, whose type is shown in Eq \ref{equ25}.

The third part is Location Loss, which is calculated based on Smooth L1 Loss \cite{fasterrcnn}:
\begin{equation}\label{equ26}
	\mathcal{L}_{loc}=\begin{cases}
		0.5\beta ^{-1}\times \left( y-\tau \right) ^2,&		\left| y-\tau \right|<\beta\\
		\left| y-\tau \right|-0.5\beta ,&		\left| y-\tau \right|\ge \beta\\
	\end{cases}
\end{equation} 
where $\beta$ is the threshold, whose value is set to 0.1 during training. $y$ and $\tau$ denote the output and real value respectively.

The last part is the Direction Loss, which is calculated based on Cross-Entropy Loss \cite{crosse}:
\begin{equation}\label{equ27}
	\mathcal{L}_{dir}=-\sum_{}^{}{P\left( \tau \right)}\log _2P\left( y \right) 
\end{equation}
where $P(\cdot)$ denotes the probability distribution of the output or real value.
 
The overall loss function is the weighted sum of all above loss items:
\begin{equation}\label{equ28}
	\mathcal{L}=\lambda _1\mathcal{L}_{cls}+\lambda _2\mathcal{L}_{occ}+\lambda _3\mathcal{L}_{loc}+\lambda _4\mathcal{L}_{dir} 
\end{equation}
where $\lambda _1$, $\lambda_2$, $\lambda _3$ and $\lambda _4$ denote the weights of loss items, they are set to 1.0, 1.0, 2.0 and 0.2 respectively during training.
\begin{table*}[t]
	\centering
	\small
	\caption{The Parameters Setting of Radar Point Clouds in Three Datasets.  \label{Hyper}}
	\renewcommand{\arraystretch}{0.55}
	\setlength{\tabcolsep}{1.8mm}{
		\begin{tabular}{cc|cc|cc|cc|cc|cc|cc|cc}
			\hline
			\multicolumn{2}{c|}{\multirow{4}[1]{*}{Datasets}} & \multicolumn{6}{c|}{\multirow{2}[1]{*}{Point Cloud Range}} & \multicolumn{2}{c|}{\multirow{4}[1]{*}{Pillar Size}} & \multicolumn{4}{c|}{\multirow{2}[1]{*}{Max Number of Pillar}} & \multicolumn{2}{c}{\multirow{4}[1]{*}{\makecell{Max Points \\Per Pillar}}} \\
			\multicolumn{2}{c|}{} & \multicolumn{6}{c|}{}                         & \multicolumn{2}{c|}{} & \multicolumn{4}{c|}{}         & \multicolumn{2}{c}{} \\
			\cline{3-8}\cline{11-14}    \multicolumn{2}{c|}{} & \multicolumn{2}{c|}{\multirow{1}[1]{*}{x}} & \multicolumn{2}{c|}{\multirow{1}[1]{*}{y}} & \multicolumn{2}{c|}{\multirow{1}[1]{*}{z}} & \multicolumn{2}{c|}{} & \multicolumn{2}{c|}{\multirow{2}[1]{*}{train }} & \multicolumn{2}{c|}{\multirow{2}[1]{*}{test}} & \multicolumn{2}{c}{} \\
			\multicolumn{2}{c|}{} & \multicolumn{2}{c|}{} & \multicolumn{2}{c|}{} & \multicolumn{2}{c|}{} & \multicolumn{2}{c|}{} & \multicolumn{2}{c|}{} & \multicolumn{2}{c|}{} & \multicolumn{2}{c}{} \\
			\hline
			\multicolumn{2}{c|}{\multirow{2}[1]{*}{VoD \cite{VoD}}} & \multicolumn{2}{c|}{\multirow{2}[1]{*}{[0.00, 51.20]}} & \multicolumn{2}{c|}{\multirow{2}[1]{*}{[-25.60,25.60]}} & \multicolumn{2}{c|}{\multirow{2}[1]{*}{[-3.00,2.00]}} & \multicolumn{2}{c|}{\multirow{2}[1]{*}{[0.16, 0.16, 0.16]}} & \multicolumn{2}{c|}{\multirow{2}[1]{*}{1.6 $\times$ 10$^4$}} & \multicolumn{2}{c|}{\multirow{2}[1]{*}{4.0 $\times$ 10$^{4}$}} & \multicolumn{2}{c}{\multirow{2}[1]{*}{5}} \\
			\multicolumn{2}{c|}{} & \multicolumn{2}{c|}{} & \multicolumn{2}{c|}{} & \multicolumn{2}{c|}{} & \multicolumn{2}{c|}{} & \multicolumn{2}{c|}{} & \multicolumn{2}{c|}{} & \multicolumn{2}{c}{} \\
			\multicolumn{2}{c|}{\multirow{2}[1]{*}{TJ4DRadSet \cite{TJ4DRadSet}}} & \multicolumn{2}{c|}{\multirow{2}[1]{*}{[0.00, 69.12]}} & \multicolumn{2}{c|}{\multirow{2}[1]{*}{[-39.68,39.68]}} & \multicolumn{2}{c|}{\multirow{2}[1]{*}{[-4.00,2.00]}} & \multicolumn{2}{c|}{\multirow{2}[1]{*}{[0.32, 0.32, 0.32]}} & \multicolumn{2}{c|}{\multirow{2}[1]{*}{1.6 $\times$ 10$^4$}} & \multicolumn{2}{c|}{\multirow{2}[1]{*}{4.0 $\times$ 10$^{4}$}} & \multicolumn{2}{c}{\multirow{2}[1]{*}{5}} \\
			\multicolumn{2}{c|}{} & \multicolumn{2}{c|}{} & \multicolumn{2}{c|}{} & \multicolumn{2}{c|}{} & \multicolumn{2}{c|}{} & \multicolumn{2}{c|}{} & \multicolumn{2}{c|}{} & \multicolumn{2}{c}{} \\
			\multicolumn{2}{c|}{\multirow{2}[1]{*}{Astyx HiRes 2019 \cite{Astyx}}} & \multicolumn{2}{c|}{\multirow{2}[1]{*}{[0.00, 76.80]}} & \multicolumn{2}{c|}{\multirow{2}[1]{*}{[-40.96,40.96]}} & \multicolumn{2}{c|}{\multirow{2}[1]{*}{[-3.00,1.00]}} & \multicolumn{2}{c|}{\multirow{2}[1]{*}{[0.16, 0.16, 4.00]}} & \multicolumn{2}{c|}{\multirow{2}[1]{*}{1.6 $\times$ 10$^4$}} & \multicolumn{2}{c|}{\multirow{2}[1]{*}{4.0 $\times$ 10$^{4}$}} & \multicolumn{2}{c}{\multirow{2}[1]{*}{5}} \\
			\multicolumn{2}{c|}{} & \multicolumn{2}{c|}{} & \multicolumn{2}{c|}{} & \multicolumn{2}{c|}{} & \multicolumn{2}{c|}{} & \multicolumn{2}{c|}{} & \multicolumn{2}{c|}{} & \multicolumn{2}{c}{} \\
			\hline
		\end{tabular}%
	}
	
\end{table*}%

\begin{table*}[htbp]
	\centering
	\small
	\caption{The Parameters Setting of Training.  \label{learning}}
	\renewcommand{\arraystretch}{0.55}
	\setlength{\tabcolsep}{3mm}{
		\begin{threeparttable}
			\begin{tabular}{cc|cc|cc|cc|cc|cc|cc}
				\hline
				\multicolumn{2}{c|}{\multirow{4}[1]{*}{Datasets}} & \multicolumn{2}{c|}{\multirow{4}[1]{*}{Epochs}} & \multicolumn{2}{c|}{\multirow{4}[1]{*}{Batch Size}} & \multicolumn{8}{c}{\multirow{2}[1]{*}{Optimizer}} \\
				\multicolumn{2}{c|}{} & \multicolumn{2}{c|}{} & \multicolumn{2}{c|}{} & \multicolumn{8}{c}{} \\
				\cline{7-14}    \multicolumn{2}{c|}{} & \multicolumn{2}{c|}{} & \multicolumn{2}{c|}{} & \multicolumn{2}{c|}{\multirow{2}[1]{*}{Type}} & \multicolumn{2}{c|}{\multirow{2}[1]{*}{Initial LR*}} & \multicolumn{2}{c|}{\multirow{2}[1]{*}{Decay Strategy}} & \multicolumn{2}{c}{\multirow{2}[1]{*}{LR* Decay}} \\
				\multicolumn{2}{c|}{} & \multicolumn{2}{c|}{} & \multicolumn{2}{c|}{} & \multicolumn{2}{c|}{} & \multicolumn{2}{c|}{} & \multicolumn{2}{c|}{} & \multicolumn{2}{c}{} \\
				\hline
				\multicolumn{2}{c|}{\multirow{2}[1]{*}{VoD \cite{VoD}}} & \multicolumn{2}{c|}{\multirow{2}[1]{*}{25}} & \multicolumn{2}{c|}{\multirow{2}[1]{*}{2}} & \multicolumn{2}{c|}{\multirow{2}[1]{*}{AdamW}} & \multicolumn{2}{c|}{\multirow{2}[1]{*}{1 $\times$ 10$^{-3}$}} & \multicolumn{2}{c|}{\multirow{2}[1]{*}{OneCycle}} & \multicolumn{2}{c}{\multirow{2}[1]{*}{0.1}} \\
				\multicolumn{2}{c|}{} & \multicolumn{2}{c|}{} & \multicolumn{2}{c|}{} & \multicolumn{2}{c|}{} & \multicolumn{2}{c|}{} & \multicolumn{2}{c|}{} & \multicolumn{2}{c}{} \\
				\multicolumn{2}{c|}{\multirow{2}[0]{*}{TJ4DRadSet \cite{TJ4DRadSet}}} & \multicolumn{2}{c|}{\multirow{2}[0]{*}{40}} & \multicolumn{2}{c|}{\multirow{2}[0]{*}{2}} & \multicolumn{2}{c|}{\multirow{2}[0]{*}{AdamW}} & \multicolumn{2}{c|}{\multirow{2}[0]{*}{1 $\times$ 10$^{-3}$}} & \multicolumn{2}{c|}{\multirow{2}[0]{*}{OneCycle}} & \multicolumn{2}{c}{\multirow{2}[0]{*}{0.1}} \\
				\multicolumn{2}{c|}{} & \multicolumn{2}{c|}{} & \multicolumn{2}{c|}{} & \multicolumn{2}{c|}{} & \multicolumn{2}{c|}{} & \multicolumn{2}{c|}{} & \multicolumn{2}{c}{} \\
				\multicolumn{2}{c|}{\multirow{2}[0]{*}{Astyx Hires 2019 \cite{Astyx}}} & \multicolumn{2}{c|}{\multirow{2}[0]{*}{120}} & \multicolumn{2}{c|}{\multirow{2}[0]{*}{2}} & \multicolumn{2}{c|}{\multirow{2}[0]{*}{AdamW}} & \multicolumn{2}{c|}{\multirow{2}[0]{*}{1 $\times$ 10$^{-4}$}} & \multicolumn{2}{c|}{\multirow{2}[0]{*}{OneCycle}} & \multicolumn{2}{c}{\multirow{2}[0]{*}{0.1}} \\
				\multicolumn{2}{c|}{} & \multicolumn{2}{c|}{} & \multicolumn{2}{c|}{} & \multicolumn{2}{c|}{} & \multicolumn{2}{c|}{} & \multicolumn{2}{c|}{} & \multicolumn{2}{c}{} \\
				\hline
			\end{tabular}
			\begin{tablenotes}
				\item { $*$ represents Learning Rate.}
			\end{tablenotes} 
		\end{threeparttable}
	}
	
\end{table*}%

\begin{table*}[t]
	\centering
	\small
	\caption{The result of comparing with state-pf-the-art methods on VoD validation set. The best results are highlighted with black boldface.\label{VoD}}
	\renewcommand{\arraystretch}{0.7}
	\setlength{\tabcolsep}{2.5mm}{
		\begin{tabular}{cc|cc|cc|cccccc|cc|cccccc|cc}
			\hline
			\multicolumn{2}{c|}{\multirow{4}[2]{*}{Method}} & \multicolumn{2}{c|}{\multirow{4}[2]{*}{Year}} & \multicolumn{2}{c|}{\multirow{4}[2]{*}{Modality}} & \multicolumn{8}{c|}{\multirow{2}[1]{*}{Entire Anotated Area}} & \multicolumn{8}{c}{\multirow{2}[1]{*}{Driving Corridor}} \\
			\multicolumn{2}{c|}{} & \multicolumn{2}{c|}{} & \multicolumn{2}{c|}{} & \multicolumn{8}{c|}{}                                         & \multicolumn{8}{c}{} \\
			\cline{7-22}    \multicolumn{2}{c|}{} & \multicolumn{2}{c|}{} & \multicolumn{2}{c|}{} & \multicolumn{2}{c}{\multirow{2}[2]{*}{Car}} & \multicolumn{2}{c}{\multirow{2}[2]{*}{Pedestrian}} & \multicolumn{2}{c|}{\multirow{2}[2]{*}{Cyclist}} & \multicolumn{2}{c|}{\multirow{2}[2]{*}{mAP}} & \multicolumn{2}{c}{\multirow{2}[2]{*}{Car}} & \multicolumn{2}{c}{\multirow{2}[2]{*}{Pedestrian}} & \multicolumn{2}{c|}{\multirow{2}[2]{*}{Cyclist}} & \multicolumn{2}{c}{\multirow{2}[2]{*}{mAP}} \\
			\multicolumn{2}{c|}{} & \multicolumn{2}{c|}{} & \multicolumn{2}{c|}{} & \multicolumn{2}{c}{} & \multicolumn{2}{c}{} & \multicolumn{2}{c|}{} & \multicolumn{2}{c|}{} & \multicolumn{2}{c}{} & \multicolumn{2}{c}{} & \multicolumn{2}{c|}{} & \multicolumn{2}{c}{} \\
			\hline
			\multicolumn{2}{c|}{\multirow{2}[1]{*}{PointPillars \cite{PointPillars}}} & \multicolumn{2}{c|}{\multirow{2}[1]{*}{2019}} & \multicolumn{2}{c|}{\multirow{2}[1]{*}{R}} & \multicolumn{2}{c}{\multirow{2}[1]{*}{37.06}} & \multicolumn{2}{c}{\multirow{2}[1]{*}{35.04}} & \multicolumn{2}{c|}{\multirow{2}[1]{*}{63.44}} & \multicolumn{2}{c|}{\multirow{2}[1]{*}{45.18}} & \multicolumn{2}{c}{\multirow{2}[1]{*}{70.15}} & \multicolumn{2}{c}{\multirow{2}[1]{*}{47.22}} & \multicolumn{2}{c|}{\multirow{2}[1]{*}{85.07}} & \multicolumn{2}{c}{\multirow{2}[1]{*}{67.48}} \\
			\multicolumn{2}{c|}{} & \multicolumn{2}{c|}{} & \multicolumn{2}{c|}{} & \multicolumn{2}{c}{} & \multicolumn{2}{c}{} & \multicolumn{2}{c|}{} & \multicolumn{2}{c|}{} & \multicolumn{2}{c}{} & \multicolumn{2}{c}{} & \multicolumn{2}{c|}{} & \multicolumn{2}{c}{} \\
			\multicolumn{2}{c|}{\multirow{2}[0]{*}{CenterPoint \cite{CenterPoint}}} & \multicolumn{2}{c|}{\multirow{2}[0]{*}{2021}} & \multicolumn{2}{c|}{\multirow{2}[0]{*}{R}} & \multicolumn{2}{c}{\multirow{2}[0]{*}{32.74}} & \multicolumn{2}{c}{\multirow{2}[0]{*}{38.00}} & \multicolumn{2}{c|}{\multirow{2}[0]{*}{65.51}} & \multicolumn{2}{c|}{\multirow{2}[0]{*}{45.42}} & \multicolumn{2}{c}{\multirow{2}[0]{*}{62.01}} & \multicolumn{2}{c}{\multirow{2}[0]{*}{48.18}} & \multicolumn{2}{c|}{\multirow{2}[0]{*}{84.98}} & \multicolumn{2}{c}{\multirow{2}[0]{*}{65.06}} \\
			\multicolumn{2}{c|}{} & \multicolumn{2}{c|}{} & \multicolumn{2}{c|}{} & \multicolumn{2}{c}{} & \multicolumn{2}{c}{} & \multicolumn{2}{c|}{} & \multicolumn{2}{c|}{} & \multicolumn{2}{c}{} & \multicolumn{2}{c}{} & \multicolumn{2}{c|}{} & \multicolumn{2}{c}{} \\
			\multicolumn{2}{c|}{\multirow{2}[0]{*}{RadarPillarNet \cite{RCFusion}}} & \multicolumn{2}{c|}{\multirow{2}[0]{*}{2023}} & \multicolumn{2}{c|}{\multirow{2}[0]{*}{R}} & \multicolumn{2}{c}{\multirow{2}[0]{*}{39.30}} & \multicolumn{2}{c}{\multirow{2}[0]{*}{35.10}} & \multicolumn{2}{c|}{\multirow{2}[0]{*}{63.63}} & \multicolumn{2}{c|}{\multirow{2}[0]{*}{46.01}} & \multicolumn{2}{c}{\multirow{2}[0]{*}{71.65}} & \multicolumn{2}{c}{\multirow{2}[0]{*}{42.80}} & \multicolumn{2}{c|}{\multirow{2}[0]{*}{83.14}} & \multicolumn{2}{c}{\multirow{2}[0]{*}{65.86}} \\
			\multicolumn{2}{c|}{} & \multicolumn{2}{c|}{} & \multicolumn{2}{c|}{} & \multicolumn{2}{c}{} & \multicolumn{2}{c}{} & \multicolumn{2}{c|}{} & \multicolumn{2}{c|}{} & \multicolumn{2}{c}{} & \multicolumn{2}{c}{} & \multicolumn{2}{c|}{} & \multicolumn{2}{c}{} \\
			\multicolumn{2}{c|}{\multirow{2}[0]{*}{SMURF \cite{SMURF}}} & \multicolumn{2}{c|}{\multirow{2}[0]{*}{2023}} & \multicolumn{2}{c|}{\multirow{2}[0]{*}{R}} & \multicolumn{2}{c}{\multirow{2}[0]{*}{42.31}} & \multicolumn{2}{c}{\multirow{2}[0]{*}{39.09}} & \multicolumn{2}{c|}{\multirow{2}[0]{*}{71.50}} & \multicolumn{2}{c|}{\multirow{2}[0]{*}{50.97}} & \multicolumn{2}{c}{\multirow{2}[0]{*}{71.74}} & \multicolumn{2}{c}{\multirow{2}[0]{*}{50.54}} & \multicolumn{2}{c|}{\multirow{2}[0]{*}{86.87}} & \multicolumn{2}{c}{\multirow{2}[0]{*}{69.72}} \\
			\multicolumn{2}{c|}{} & \multicolumn{2}{c|}{} & \multicolumn{2}{c|}{} & \multicolumn{2}{c}{} & \multicolumn{2}{c}{} & \multicolumn{2}{c|}{} & \multicolumn{2}{c|}{} & \multicolumn{2}{c}{} & \multicolumn{2}{c}{} & \multicolumn{2}{c|}{} & \multicolumn{2}{c}{} \\
			\multicolumn{2}{c|}{\multirow{2}[0]{*}{MUFASA \cite{MUFASA}}} & \multicolumn{2}{c|}{\multirow{2}[0]{*}{2024}} & \multicolumn{2}{c|}{\multirow{2}[0]{*}{R}} & \multicolumn{2}{c}{\multirow{2}[0]{*}{43.10}} & \multicolumn{2}{c}{\multirow{2}[0]{*}{38.97}} & \multicolumn{2}{c|}{\multirow{2}[0]{*}{68.65}} & \multicolumn{2}{c|}{\multirow{2}[0]{*}{50.24}} & \multicolumn{2}{c}{\multirow{2}[0]{*}{72.50}} & \multicolumn{2}{c}{\multirow{2}[0]{*}{50.28}} & \multicolumn{2}{c|}{\multirow{2}[0]{*}{88.51}} & \multicolumn{2}{c}{\multirow{2}[0]{*}{70.43}} \\
			\multicolumn{2}{c|}{} & \multicolumn{2}{c|}{} & \multicolumn{2}{c|}{} & \multicolumn{2}{c}{} & \multicolumn{2}{c}{} & \multicolumn{2}{c|}{} & \multicolumn{2}{c|}{} & \multicolumn{2}{c}{} & \multicolumn{2}{c}{} & \multicolumn{2}{c|}{} & \multicolumn{2}{c}{} \\
			\multicolumn{2}{c|}{\multirow{2}[0]{*}{Ruddat et al. \cite{ruddat}}} & \multicolumn{2}{c|}{\multirow{2}[0]{*}{2024}} & \multicolumn{2}{c|}{\multirow{2}[0]{*}{R}} & \multicolumn{2}{c}{\multirow{2}[0]{*}{36.70}} & \multicolumn{2}{c}{\multirow{2}[0]{*}{36.80}} & \multicolumn{2}{c|}{\multirow{2}[0]{*}{65.00}} & \multicolumn{2}{c|}{\multirow{2}[0]{*}{46.20}} & \multicolumn{2}{c}{\multirow{2}[0]{*}{69.10}} & \multicolumn{2}{c}{\multirow{2}[0]{*}{47.20}} & \multicolumn{2}{c|}{\multirow{2}[0]{*}{84.30}} & \multicolumn{2}{c}{\multirow{2}[0]{*}{66.90}} \\
			\multicolumn{2}{c|}{} & \multicolumn{2}{c|}{} & \multicolumn{2}{c|}{} & \multicolumn{2}{c}{} & \multicolumn{2}{c}{} & \multicolumn{2}{c|}{} & \multicolumn{2}{c|}{} & \multicolumn{2}{c}{} & \multicolumn{2}{c}{} & \multicolumn{2}{c|}{} & \multicolumn{2}{c}{} \\
			\multicolumn{2}{c|}{\multirow{2}[0]{*}{Duan et al. \cite{duan}}} & \multicolumn{2}{c|}{\multirow{2}[0]{*}{2025}} & \multicolumn{2}{c|}{\multirow{2}[0]{*}{R}} & \multicolumn{2}{c}{\multirow{2}[0]{*}{39.18}} & \multicolumn{2}{c}{\multirow{2}[0]{*}{37.23}} & \multicolumn{2}{c|}{\multirow{2}[0]{*}{66.81}} & \multicolumn{2}{c|}{\multirow{2}[0]{*}{47.74}} & \multicolumn{2}{c}{\multirow{2}[0]{*}{71.87}} & \multicolumn{2}{c}{\multirow{2}[0]{*}{49.45}} & \multicolumn{2}{c|}{\multirow{2}[0]{*}{86.43}} & \multicolumn{2}{c}{\multirow{2}[0]{*}{69.25}} \\
			\multicolumn{2}{c|}{} & \multicolumn{2}{c|}{} & \multicolumn{2}{c|}{} & \multicolumn{2}{c}{} & \multicolumn{2}{c}{} & \multicolumn{2}{c|}{} & \multicolumn{2}{c|}{} & \multicolumn{2}{c}{} & \multicolumn{2}{c}{} & \multicolumn{2}{c|}{} & \multicolumn{2}{c}{} \\
			\multicolumn{2}{c|}{\multirow{2}[0]{*}{MAFF-Net \cite{MAFF-Net}}} & \multicolumn{2}{c|}{\multirow{2}[0]{*}{2025}} & \multicolumn{2}{c|}{\multirow{2}[0]{*}{R}} & \multicolumn{2}{c}{\multirow{2}[0]{*}{42.33}} & \multicolumn{2}{c}{\multirow{2}[0]{*}{46.75}} & \multicolumn{2}{c|}{\multirow{2}[0]{*}{74.72}} & \multicolumn{2}{c|}{\multirow{2}[0]{*}{54.59}} & \multicolumn{2}{c}{\multirow{2}[0]{*}{72.78}} & \multicolumn{2}{c}{\multirow{2}[0]{*}{57.81}} & \multicolumn{2}{c|}{\multirow{2}[0]{*}{87.40}} & \multicolumn{2}{c}{\multirow{2}[0]{*}{72.50}} \\
			\multicolumn{2}{c|}{} & \multicolumn{2}{c|}{} & \multicolumn{2}{c|}{} & \multicolumn{2}{c}{} & \multicolumn{2}{c}{} & \multicolumn{2}{c|}{} & \multicolumn{2}{c|}{} & \multicolumn{2}{c}{} & \multicolumn{2}{c}{} & \multicolumn{2}{c|}{} & \multicolumn{2}{c}{} \\
			\hline
			\multicolumn{2}{c|}{\multirow{2}[1]{*}{PointAugmenting \cite{pointaugmenting}}} & \multicolumn{2}{c|}{\multirow{2}[1]{*}{2021}} & \multicolumn{2}{c|}{\multirow{2}[1]{*}{R+C}} & \multicolumn{2}{c}{\multirow{2}[1]{*}{39.62}} & \multicolumn{2}{c}{\multirow{2}[1]{*}{44.48}} & \multicolumn{2}{c|}{\multirow{2}[1]{*}{73.70}} & \multicolumn{2}{c|}{\multirow{2}[1]{*}{52.60}} & \multicolumn{2}{c}{\multirow{2}[1]{*}{71.02}} & \multicolumn{2}{c}{\multirow{2}[1]{*}{48.59}} & \multicolumn{2}{c|}{\multirow{2}[1]{*}{87.57}} & \multicolumn{2}{c}{\multirow{2}[1]{*}{69.06}} \\
			\multicolumn{2}{c|}{} & \multicolumn{2}{c|}{} & \multicolumn{2}{c|}{} & \multicolumn{2}{c}{} & \multicolumn{2}{c}{} & \multicolumn{2}{c|}{} & \multicolumn{2}{c|}{} & \multicolumn{2}{c}{} & \multicolumn{2}{c}{} & \multicolumn{2}{c|}{} & \multicolumn{2}{c}{} \\
			\multicolumn{2}{c|}{\multirow{2}[0]{*}{RCFusion \cite{RCFusion}}} & \multicolumn{2}{c|}{\multirow{2}[0]{*}{2023}} & \multicolumn{2}{c|}{\multirow{2}[0]{*}{R+C}} & \multicolumn{2}{c}{\multirow{2}[0]{*}{41.70}} & \multicolumn{2}{c}{\multirow{2}[0]{*}{38.95}} & \multicolumn{2}{c|}{\multirow{2}[0]{*}{68.31}} & \multicolumn{2}{c|}{\multirow{2}[0]{*}{49.65}} & \multicolumn{2}{c}{\multirow{2}[0]{*}{71.87}} & \multicolumn{2}{c}{\multirow{2}[0]{*}{47.50}} & \multicolumn{2}{c|}{\multirow{2}[0]{*}{88.33}} & \multicolumn{2}{c}{\multirow{2}[0]{*}{69.23}} \\
			\multicolumn{2}{c|}{} & \multicolumn{2}{c|}{} & \multicolumn{2}{c|}{} & \multicolumn{2}{c}{} & \multicolumn{2}{c}{} & \multicolumn{2}{c|}{} & \multicolumn{2}{c|}{} & \multicolumn{2}{c}{} & \multicolumn{2}{c}{} & \multicolumn{2}{c|}{} & \multicolumn{2}{c}{} \\
			\multicolumn{2}{c|}{\multirow{2}[0]{*}{FUTR3D \cite{FUTR3D}}} & \multicolumn{2}{c|}{\multirow{2}[0]{*}{2023}} & \multicolumn{2}{c|}{\multirow{2}[0]{*}{R+C}} & \multicolumn{2}{c}{\multirow{2}[0]{*}{46.01}} & \multicolumn{2}{c}{\multirow{2}[0]{*}{35.11}} & \multicolumn{2}{c|}{\multirow{2}[0]{*}{65.98}} & \multicolumn{2}{c|}{\multirow{2}[0]{*}{49.03}} & \multicolumn{2}{c}{\multirow{2}[0]{*}{78.66}} & \multicolumn{2}{c}{\multirow{2}[0]{*}{43.10}} & \multicolumn{2}{c|}{\multirow{2}[0]{*}{86.19}} & \multicolumn{2}{c}{\multirow{2}[0]{*}{69.32}} \\
			\multicolumn{2}{c|}{} & \multicolumn{2}{c|}{} & \multicolumn{2}{c|}{} & \multicolumn{2}{c}{} & \multicolumn{2}{c}{} & \multicolumn{2}{c|}{} & \multicolumn{2}{c|}{} & \multicolumn{2}{c}{} & \multicolumn{2}{c}{} & \multicolumn{2}{c|}{} & \multicolumn{2}{c}{} \\
			\multicolumn{2}{c|}{\multirow{2}[0]{*}{BEVFusion \cite{BEVFusion}}} & \multicolumn{2}{c|}{\multirow{2}[0]{*}{2023}} & \multicolumn{2}{c|}{\multirow{2}[0]{*}{R+C}} & \multicolumn{2}{c}{\multirow{2}[0]{*}{37.85}} & \multicolumn{2}{c}{\multirow{2}[0]{*}{40.96}} & \multicolumn{2}{c|}{\multirow{2}[0]{*}{68.95}} & \multicolumn{2}{c|}{\multirow{2}[0]{*}{49.25}} & \multicolumn{2}{c}{\multirow{2}[0]{*}{70.21}} & \multicolumn{2}{c}{\multirow{2}[0]{*}{45.86}} & \multicolumn{2}{c|}{\multirow{2}[0]{*}{89.48}} & \multicolumn{2}{c}{\multirow{2}[0]{*}{68.52}} \\
			\multicolumn{2}{c|}{} & \multicolumn{2}{c|}{} & \multicolumn{2}{c|}{} & \multicolumn{2}{c}{} & \multicolumn{2}{c}{} & \multicolumn{2}{c|}{} & \multicolumn{2}{c|}{} & \multicolumn{2}{c}{} & \multicolumn{2}{c}{} & \multicolumn{2}{c|}{} & \multicolumn{2}{c}{} \\
			\multicolumn{2}{c|}{\multirow{2}[0]{*}{LXL \cite{LxL}}} & \multicolumn{2}{c|}{\multirow{2}[0]{*}{2023}} & \multicolumn{2}{c|}{\multirow{2}[0]{*}{R+C}} & \multicolumn{2}{c}{\multirow{2}[0]{*}{42.33}} & \multicolumn{2}{c}{\multirow{2}[0]{*}{49.48}} & \multicolumn{2}{c|}{\multirow{2}[0]{*}{\textbf{77.12}}} & \multicolumn{2}{c|}{\multirow{2}[0]{*}{56.31}} & \multicolumn{2}{c}{\multirow{2}[0]{*}{72.18}} & \multicolumn{2}{c}{\multirow{2}[0]{*}{58.30}} & \multicolumn{2}{c|}{\multirow{2}[0]{*}{88.31}} & \multicolumn{2}{c}{\multirow{2}[0]{*}{72.93}} \\
			\multicolumn{2}{c|}{} & \multicolumn{2}{c|}{} & \multicolumn{2}{c|}{} & \multicolumn{2}{c}{} & \multicolumn{2}{c}{} & \multicolumn{2}{c|}{} & \multicolumn{2}{c|}{} & \multicolumn{2}{c}{} & \multicolumn{2}{c}{} & \multicolumn{2}{c|}{} & \multicolumn{2}{c}{} \\
			\multicolumn{2}{c|}{\multirow{2}[0]{*}{RCBEVDet \cite{RCBEVDet}}} & \multicolumn{2}{c|}{\multirow{2}[0]{*}{2024}} & \multicolumn{2}{c|}{\multirow{2}[0]{*}{R+C}} & \multicolumn{2}{c}{\multirow{2}[0]{*}{40.63}} & \multicolumn{2}{c}{\multirow{2}[0]{*}{38.86}} & \multicolumn{2}{c|}{\multirow{2}[0]{*}{70.48}} & \multicolumn{2}{c|}{\multirow{2}[0]{*}{49.99}} & \multicolumn{2}{c}{\multirow{2}[0]{*}{72.48}} & \multicolumn{2}{c}{\multirow{2}[0]{*}{49.89}} & \multicolumn{2}{c|}{\multirow{2}[0]{*}{87.01}} & \multicolumn{2}{c}{\multirow{2}[0]{*}{69.80}} \\
			\multicolumn{2}{c|}{} & \multicolumn{2}{c|}{} & \multicolumn{2}{c|}{} & \multicolumn{2}{c}{} & \multicolumn{2}{c}{} & \multicolumn{2}{c|}{} & \multicolumn{2}{c|}{} & \multicolumn{2}{c}{} & \multicolumn{2}{c}{} & \multicolumn{2}{c|}{} & \multicolumn{2}{c}{} \\
			\multicolumn{2}{c|}{\multirow{2}[0]{*}{IS-Fusion \cite{IS-Fusion}}} & \multicolumn{2}{c|}{\multirow{2}[0]{*}{2024}} & \multicolumn{2}{c|}{\multirow{2}[0]{*}{R+C}} & \multicolumn{2}{c}{\multirow{2}[0]{*}{48.57}} & \multicolumn{2}{c}{\multirow{2}[0]{*}{46.17}} & \multicolumn{2}{c|}{\multirow{2}[0]{*}{68.48}} & \multicolumn{2}{c|}{\multirow{2}[0]{*}{54.40}} & \multicolumn{2}{c}{\multirow{2}[0]{*}{80.42}} & \multicolumn{2}{c}{\multirow{2}[0]{*}{55.50}} & \multicolumn{2}{c|}{\multirow{2}[0]{*}{88.33}} & \multicolumn{2}{c}{\multirow{2}[0]{*}{74.75}} \\
			\multicolumn{2}{c|}{} & \multicolumn{2}{c|}{} & \multicolumn{2}{c|}{} & \multicolumn{2}{c}{} & \multicolumn{2}{c}{} & \multicolumn{2}{c|}{} & \multicolumn{2}{c|}{} & \multicolumn{2}{c}{} & \multicolumn{2}{c}{} & \multicolumn{2}{c|}{} & \multicolumn{2}{c}{} \\
			\multicolumn{2}{c|}{\multirow{2}[0]{*}{TL-4DRCF \cite{TL-4DRCF}}} & \multicolumn{2}{c|}{\multirow{2}[0]{*}{2024}} & \multicolumn{2}{c|}{\multirow{2}[0]{*}{R+C}} & \multicolumn{2}{c}{\multirow{2}[0]{*}{43.71}} & \multicolumn{2}{c}{\multirow{2}[0]{*}{40.11}} & \multicolumn{2}{c|}{\multirow{2}[0]{*}{64.22}} & \multicolumn{2}{c|}{\multirow{2}[0]{*}{49.35}} & \multicolumn{2}{c}{\multirow{2}[0]{*}{79.49}} & \multicolumn{2}{c}{\multirow{2}[0]{*}{53.76}} & \multicolumn{2}{c|}{\multirow{2}[0]{*}{76.50}} & \multicolumn{2}{c}{\multirow{2}[0]{*}{69.92}} \\
			\multicolumn{2}{c|}{} & \multicolumn{2}{c|}{} & \multicolumn{2}{c|}{} & \multicolumn{2}{c}{} & \multicolumn{2}{c}{} & \multicolumn{2}{c|}{} & \multicolumn{2}{c|}{} & \multicolumn{2}{c}{} & \multicolumn{2}{c}{} & \multicolumn{2}{c|}{} & \multicolumn{2}{c}{} \\
			\multicolumn{2}{c|}{\multirow{2}[0]{*}{RCDFNet \cite{RCDFNet}}} & \multicolumn{2}{c|}{\multirow{2}[0]{*}{2025}} & \multicolumn{2}{c|}{\multirow{2}[0]{*}{R+C}} & \multicolumn{2}{c}{\multirow{2}[0]{*}{48.65}} & \multicolumn{2}{c}{\multirow{2}[0]{*}{45.27}} & \multicolumn{2}{c|}{\multirow{2}[0]{*}{76.06}} & \multicolumn{2}{c|}{\multirow{2}[0]{*}{56.66}} & \multicolumn{2}{c}{\multirow{2}[0]{*}{72.46}} & \multicolumn{2}{c}{\multirow{2}[0]{*}{49.76}} & \multicolumn{2}{c|}{\multirow{2}[0]{*}{89.62}} & \multicolumn{2}{c}{\multirow{2}[0]{*}{70.61}} \\
			\multicolumn{2}{c|}{} & \multicolumn{2}{c|}{} & \multicolumn{2}{c|}{} & \multicolumn{2}{c}{} & \multicolumn{2}{c}{} & \multicolumn{2}{c|}{} & \multicolumn{2}{c|}{} & \multicolumn{2}{c}{} & \multicolumn{2}{c}{} & \multicolumn{2}{c|}{} & \multicolumn{2}{c}{} \\
			\multicolumn{2}{c|}{\multirow{2}[0]{*}{Wang et al. \cite{Wang}}} & \multicolumn{2}{c|}{\multirow{2}[0]{*}{2025}} & \multicolumn{2}{c|}{\multirow{2}[0]{*}{R+C}} & \multicolumn{2}{c}{\multirow{2}[0]{*}{42.24}} & \multicolumn{2}{c}{\multirow{2}[0]{*}{40.38}} & \multicolumn{2}{c|}{\multirow{2}[0]{*}{69.12}} & \multicolumn{2}{c|}{\multirow{2}[0]{*}{50.58}} & \multicolumn{2}{c}{\multirow{2}[0]{*}{72.35}} & \multicolumn{2}{c}{\multirow{2}[0]{*}{54.10}} & \multicolumn{2}{c|}{\multirow{2}[0]{*}{88.78}} & \multicolumn{2}{c}{\multirow{2}[0]{*}{71.74}} \\
			\multicolumn{2}{c|}{} & \multicolumn{2}{c|}{} & \multicolumn{2}{c|}{} & \multicolumn{2}{c}{} & \multicolumn{2}{c}{} & \multicolumn{2}{c|}{} & \multicolumn{2}{c|}{} & \multicolumn{2}{c}{} & \multicolumn{2}{c}{} & \multicolumn{2}{c|}{} & \multicolumn{2}{c}{} \\
			\multicolumn{2}{c|}{\multirow{2}[0]{*}{Li et al. \cite{Li}}} & \multicolumn{2}{c|}{\multirow{2}[0]{*}{2025}} & \multicolumn{2}{c|}{\multirow{2}[0]{*}{R+C}} & \multicolumn{2}{c}{\multirow{2}[0]{*}{52.11}} & \multicolumn{2}{c}{\multirow{2}[0]{*}{43.58}} & \multicolumn{2}{c|}{\multirow{2}[0]{*}{59.45}} & \multicolumn{2}{c|}{\multirow{2}[0]{*}{51.72}} & \multicolumn{2}{c}{\multirow{2}[0]{*}{88.86}} & \multicolumn{2}{c}{\multirow{2}[0]{*}{62.43}} & \multicolumn{2}{c|}{\multirow{2}[0]{*}{87.07}} & \multicolumn{2}{c}{\multirow{2}[0]{*}{79.45}} \\
			\multicolumn{2}{c|}{} & \multicolumn{2}{c|}{} & \multicolumn{2}{c|}{} & \multicolumn{2}{c}{} & \multicolumn{2}{c}{} & \multicolumn{2}{c|}{} & \multicolumn{2}{c|}{} & \multicolumn{2}{c}{} & \multicolumn{2}{c}{} & \multicolumn{2}{c|}{} & \multicolumn{2}{c}{} \\
			\multicolumn{2}{c|}{\multirow{2}[0]{*}{UniBEVFusion \cite{UniBEVFusion}}} & \multicolumn{2}{c|}{\multirow{2}[0]{*}{2025}} & \multicolumn{2}{c|}{\multirow{2}[0]{*}{R+C}} & \multicolumn{2}{c}{\multirow{2}[0]{*}{42.22}} & \multicolumn{2}{c}{\multirow{2}[0]{*}{47.11}} & \multicolumn{2}{c|}{\multirow{2}[0]{*}{72.94}} & \multicolumn{2}{c|}{\multirow{2}[0]{*}{54.09}} & \multicolumn{2}{c}{\multirow{2}[0]{*}{72.10}} & \multicolumn{2}{c}{\multirow{2}[0]{*}{57.71}} & \multicolumn{2}{c|}{\multirow{2}[0]{*}{\textbf{93.29}}} & \multicolumn{2}{c}{\multirow{2}[0]{*}{74.37}} \\
			\multicolumn{2}{c|}{} & \multicolumn{2}{c|}{} & \multicolumn{2}{c|}{} & \multicolumn{2}{c}{} & \multicolumn{2}{c}{} & \multicolumn{2}{c|}{} & \multicolumn{2}{c|}{} & \multicolumn{2}{c}{} & \multicolumn{2}{c}{} & \multicolumn{2}{c|}{} & \multicolumn{2}{c}{} \\
			\multicolumn{2}{c|}{\multirow{2}[0]{*}{ZFusion \cite{ZFusion}}} & \multicolumn{2}{c|}{\multirow{2}[0]{*}{2025}} & \multicolumn{2}{c|}{\multirow{2}[0]{*}{R+C}} & \multicolumn{2}{c}{\multirow{2}[0]{*}{44.77}} & \multicolumn{2}{c}{\multirow{2}[0]{*}{40.02}} & \multicolumn{2}{c|}{\multirow{2}[0]{*}{68.61}} & \multicolumn{2}{c|}{\multirow{2}[0]{*}{51.14}} & \multicolumn{2}{c}{\multirow{2}[0]{*}{80.44}} & \multicolumn{2}{c}{\multirow{2}[0]{*}{52.68}} & \multicolumn{2}{c|}{\multirow{2}[0]{*}{90.01}} & \multicolumn{2}{c}{\multirow{2}[0]{*}{74.38}} \\
			\multicolumn{2}{c|}{} & \multicolumn{2}{c|}{} & \multicolumn{2}{c|}{} & \multicolumn{2}{c}{} & \multicolumn{2}{c}{} & \multicolumn{2}{c|}{} & \multicolumn{2}{c|}{} & \multicolumn{2}{c}{} & \multicolumn{2}{c}{} & \multicolumn{2}{c|}{} & \multicolumn{2}{c}{} \\
			\multicolumn{2}{c|}{\multirow{2}[0]{*}{SFGFusion \cite{SFGFusion}}} & \multicolumn{2}{c|}{\multirow{2}[0]{*}{2025}} & \multicolumn{2}{c|}{\multirow{2}[0]{*}{R+C}} & \multicolumn{2}{c}{\multirow{2}[0]{*}{48.30}} & \multicolumn{2}{c}{\multirow{2}[0]{*}{43.60}} & \multicolumn{2}{c|}{\multirow{2}[0]{*}{75.54}} & \multicolumn{2}{c|}{\multirow{2}[0]{*}{55.76}} & \multicolumn{2}{c}{\multirow{2}[0]{*}{79.10}} & \multicolumn{2}{c}{\multirow{2}[0]{*}{53.63}} & \multicolumn{2}{c|}{\multirow{2}[0]{*}{88.60}} & \multicolumn{2}{c}{\multirow{2}[0]{*}{73.77}} \\
			\multicolumn{2}{c|}{} & \multicolumn{2}{c|}{} & \multicolumn{2}{c|}{} & \multicolumn{2}{c}{} & \multicolumn{2}{c}{} & \multicolumn{2}{c|}{} & \multicolumn{2}{c|}{} & \multicolumn{2}{c}{} & \multicolumn{2}{c}{} & \multicolumn{2}{c|}{} & \multicolumn{2}{c}{} \\
			\multicolumn{2}{c|}{\multirow{2}[0]{*}{SGDet3D \cite{SGDet3D}}} & \multicolumn{2}{c|}{\multirow{2}[0]{*}{2025}} & \multicolumn{2}{c|}{\multirow{2}[0]{*}{R+C}} & \multicolumn{2}{c}{\multirow{2}[0]{*}{53.16}} & \multicolumn{2}{c}{\multirow{2}[0]{*}{49.98}} & \multicolumn{2}{c|}{\multirow{2}[0]{*}{76.11}} & \multicolumn{2}{c|}{\multirow{2}[0]{*}{59.75}} & \multicolumn{2}{c}{\multirow{2}[0]{*}{81.13}} & \multicolumn{2}{c}{\multirow{2}[0]{*}{60.91}} & \multicolumn{2}{c|}{\multirow{2}[0]{*}{90.22}} & \multicolumn{2}{c}{\multirow{2}[0]{*}{77.42}} \\
			\multicolumn{2}{c|}{} & \multicolumn{2}{c|}{} & \multicolumn{2}{c|}{} & \multicolumn{2}{c}{} & \multicolumn{2}{c}{} & \multicolumn{2}{c|}{} & \multicolumn{2}{c|}{} & \multicolumn{2}{c}{} & \multicolumn{2}{c}{} & \multicolumn{2}{c|}{} & \multicolumn{2}{c}{} \\
			\multicolumn{2}{c|}{\multirow{2}[0]{*}{DSFusion \cite{DSFusion}}} & \multicolumn{2}{c|}{\multirow{2}[0]{*}{2025}} & \multicolumn{2}{c|}{\multirow{2}[0]{*}{R+C}} & \multicolumn{2}{c}{\multirow{2}[0]{*}{45.75}} & \multicolumn{2}{c}{\multirow{2}[0]{*}{50.71}} & \multicolumn{2}{c|}{\multirow{2}[0]{*}{72.64}} & \multicolumn{2}{c|}{\multirow{2}[0]{*}{56.37}} & \multicolumn{2}{c}{\multirow{2}[0]{*}{78.21}} & \multicolumn{2}{c}{\multirow{2}[0]{*}{61.22}} & \multicolumn{2}{c|}{\multirow{2}[0]{*}{87.73}} & \multicolumn{2}{c}{\multirow{2}[0]{*}{75.72}} \\
			\multicolumn{2}{c|}{} & \multicolumn{2}{c|}{} & \multicolumn{2}{c|}{} & \multicolumn{2}{c}{} & \multicolumn{2}{c}{} & \multicolumn{2}{c|}{} & \multicolumn{2}{c|}{} & \multicolumn{2}{c}{} & \multicolumn{2}{c}{} & \multicolumn{2}{c|}{} & \multicolumn{2}{c}{} \\
			\multicolumn{2}{c|}{\multirow{2}[0]{*}{HSGFusion \cite{HGSFusion}}} & \multicolumn{2}{c|}{\multirow{2}[0]{*}{2025}} & \multicolumn{2}{c|}{\multirow{2}[0]{*}{R+C}} & \multicolumn{2}{c}{\multirow{2}[0]{*}{51.67}} & \multicolumn{2}{c}{\multirow{2}[0]{*}{52.39}} & \multicolumn{2}{c|}{\multirow{2}[0]{*}{68.50}} & \multicolumn{2}{c|}{\multirow{2}[0]{*}{57.72}} & \multicolumn{2}{c}{\multirow{2}[0]{*}{87.24}} & \multicolumn{2}{c}{\multirow{2}[0]{*}{58.66}} & \multicolumn{2}{c|}{\multirow{2}[0]{*}{87.49}} & \multicolumn{2}{c}{\multirow{2}[0]{*}{77.79}} \\
			\multicolumn{2}{c|}{} & \multicolumn{2}{c|}{} & \multicolumn{2}{c|}{} & \multicolumn{2}{c}{} & \multicolumn{2}{c}{} & \multicolumn{2}{c|}{} & \multicolumn{2}{c|}{} & \multicolumn{2}{c}{} & \multicolumn{2}{c}{} & \multicolumn{2}{c|}{} & \multicolumn{2}{c}{} \\
			\multicolumn{2}{c|}{\multirow{2}[0]{*}{Ours}} & \multicolumn{2}{c|}{\multirow{2}[0]{*}{-}} & \multicolumn{2}{c|}{\multirow{2}[0]{*}{R+C}} & \multicolumn{2}{c}{\multirow{2}[0]{*}{\textbf{55.28}}} & \multicolumn{2}{c}{\multirow{2}[0]{*}{\textbf{53.75}}} & \multicolumn{2}{c|}{\multirow{2}[0]{*}{71.04}} & \multicolumn{2}{c|}{\multirow{2}[0]{*}{\textbf{60.02}}} & \multicolumn{2}{c}{\multirow{2}[0]{*}{\textbf{89.07}}} & \multicolumn{2}{c}{\multirow{2}[0]{*}{\textbf{67.04}}} & \multicolumn{2}{c|}{\multirow{2}[0]{*}{89.24}} & \multicolumn{2}{c}{\multirow{2}[0]{*}{\textbf{81.78}}} \\
			\multicolumn{2}{c|}{} & \multicolumn{2}{c|}{} & \multicolumn{2}{c|}{} & \multicolumn{2}{c}{} & \multicolumn{2}{c}{} & \multicolumn{2}{c|}{} & \multicolumn{2}{c|}{} & \multicolumn{2}{c}{} & \multicolumn{2}{c}{} & \multicolumn{2}{c|}{} & \multicolumn{2}{c}{} \\
			\hline
	\end{tabular}}
	
\end{table*}%
\section{Experiment}
\subsection{Implementation Details}
\subsubsection{Dataset and Metrics}
The datasets chosen for evaluating the performance of proposed method are View-of-Delft \cite{VoD} (VoD), TJ4DRadSet \cite{TJ4DRadSet} and Astyx HiRes 2019 \cite{Astyx} dataset.

The VoD dataset is a new autonomous driving dataset which contains 64-layer LIDAR, 3+1-D radar and stereo camera datas. The source of datas came from Delft mainly. There are three types of radar point clouds: single-scan, three-scan and five-scan. The five-scan point cloud data is adopted to detect three catagories, which include Car, Pedestrian and Cyclist. The metrics provided by VoD dataset is 3-D mean average precision (mAP) in the entire annotated area (EAA) and the driving corridor (DC).

The TJ4DRadSet dataset is another novel autonomous driving dataset whose source of datas came from Singapore and Suzhou, China mainly. The data contains 3+1-D radar point clouds in
continuous sequences with 3-D annotations, and 
multimodal complete information from 64-layers LIDAR, camera, and GNSS. The scenes includes daylight, night, and exposure. The metrics adopted by TJ4DRadSet dataset are 3-D mAP and BEV mAP, with detected instances contains Car, Pedestrian, Cyclist and Truck. 

The Astyx HiRes 2019 dataset is a small autonomous driving dataset which contains 64-layers LIDAR, 3+1-D radar and stereo camera datas captured in Germany urban and outskirt roads. The dataset contains three object categories, including Car, Pedestrian, and Cyclist, with detailed 3-D bounding box annotations which indicate the sensor modality responsible for detecting each object. The evaluation metrics employ 3-D mAP and BEV mAP in three difficulty levels which include Easy, Moderate and Hard.
 
The experiment adopts the metrics provided by official datasets respectively. For all dataset, the  intersection-over-union (IoU) thresholds are set 0.25 for Pedestrian and Cyclist, and 0.5 for Car and Truck. 

\subsubsection{Training Details}
The ratio of training and testing set is set to 4:1 for VoD and Astyx HiRes 2019 dataset, while 5:1 for TJ4DRadSet dataset. Moreover, the parameters setting of radar point clouds in three datasets is shown in Table \ref{Hyper}. All of the training are conducted on eight NVIDIA RTX 3090 GPUs within OpenPCDet \cite{openpcdet} Platform, with Python 3.8, PyTorch 2.0.0 and CUDA 11.8 under Ubuntu 20.04. During inference, the data type of tensors on the GPUs is Half-Precision Floating-Point (FP16). The parameters setting of training is shown in Table \ref{learning}. Most of the experiment results are visualized through MATLAB 2020.

For SimDen module, YOLOv8-Seg \cite{YOLOv8} large model is adopted for the inference of instance segmetation. In addition, Gauss kernel function and Silverman \cite{silverman} bandwidth rule are adopted for 3-D KDE program. The 2-D points own top 4 high density value are chosen for Gaussian simulation. The number of edge points adopted for curvature estimation is 20. The number of simulated points on every instances is 200.

For RCM and MMIF module, the size of input feature maps is 320 $\times$ 320. In addition, the number of convolution layers is 2, and they contain 3 different receptive fields.
The channel transformation in MMIF module is Reshape operator, which transforms the size of input feature maps to 80 $\times$ 80, improving the number channels from 128 to 128 $\times$ 16. The number of DWConv2D-SiLU layers is 2 for feature enhancement.

The backbones of radar and vision datas are RadarPillarNet \cite{RCFusion} and VMamba \cite{Vmamba} respectively. In addition, similar to Faster-RCNN \cite{fasterrcnn}, the Region Proposal Network (RPN) artchitecture is adopted as the detection head of SDCM. 

\subsection{Comparing with State-of-the-art Methods}
\subsubsection{On the VoD Validation Set}
Table \ref{VoD} shows the comparison results on the VoD validation set. Our method outperforms many single-modality approaches, such as MUFASA \cite{MUFASA} and Ruddat et al. \cite{ruddat} in 2024, Duan et al. \cite{duan} and MAFF-Net \cite{MAFF-Net} in 2025. especially, our method improves about 5.43 points in EAA mAP and about 9.28 points in DC mAP, comparing with MAFF-Net which performs the best in single-modality methods.

Comparing with multi-modality approaches, our methods also outperform them, especially for 2025 approaches such as RCDFNet \cite{RCDFNet}, Wang et al. \cite{Wang}, Li et al. \cite{Li}, UniBEVFusion \cite{UniBEVFusion}, ZFusion \cite{ZFusion}, SFGFusion \cite{SFGFusion}, SGDet3D \cite{SGDet3D}, DSFusion \cite{DSFusion} and HGSFusion \cite{HGSFusion}. Comapring with LXL \cite{LxL} which perform the best before 2025, our method improves about 3.71 points in EAA mAP and 8.85 points in DC mAP. Moreover, our method improves about 0.27 and 2.30 points in EAA mAP, as well as about 4.36 and 3.99 points in DC mAP respectively, comparing with SGDet3D and HGSFusion. 

However, our method has no abvious advantages on the detection of Cyclist category, mainly because SimDen module focuses more on the point cloud densifying of persons on the bicycles or motorcycles instead of the entirety, which brings false detection. Nonetheless, for Cyclist category, our method also outperforms novel approaches such as ZFusion, HGSFusion and so on.

\subsubsection{On the TJ4DRadSet Test Set}
Table \ref{tj4d} shows the the comparison results on the TJ4DRadSet test set. Our method outperform single-modality approaches on this dataset. Comparing with MAFF-Net \cite{MAFF-Net}, our method improves about 7.67 points in 3-D mAP and about 7.74 points in BEV mAP. For multi-modality approaches, our method 
improves about 5.12 points in 3-D mAP and 11.87 points in BEV mAP, comparing with RCDFNet. Our method also outperforms SGDet3D with about 1.23 and 2.17 points in 3-D and BEV mAP respectively.

It cloud be seen that the Cyclist category is detected well by our method on TJ4DRadSet dataset, mainly because the Cyclists with smaller semantic masks than VoD dataset, which ensures that the point cloud densifying focuses on the entirety. For Truck category, its semantic masks' areas are larger than other categories, especially in the short range. Therefore, the 3-D KDE values of their point clouds are close to uniform, leading to blank space on the point sets after Gaussian simulation. This is reason why our method has no abvious advantages on the detection of Truck. 

\subsubsection{On the Astyx HiRes 2019 Test set}
Table \ref{Astyx} shows the comparison results on the Astyx HiRes 2019 test set. The comparison focuses on three difficulty levels on 3-D and BEV mAP. Our method outperforms 2025 single-modality approaches like Duan et al. with about 20.71/5.65/3.51 points improvements in 3-D mAP and 5.09/5.42/2.40 points improvements in BEV mAP. Especially, our method outperforms HGSFusion with about 4.45/0.63/2.67 points imrpovement in 3-D mAP and 1.27/1.75/0.09 points improvements in BEV mAP. This experiment shows that our method have good generalization ability.
\begin{table*}[t]
	\centering
	\small
	\caption{The result of comparing with state-pf-the-art methods on TJ4DRadSet test set. The best results are highlighted with black boldface. 	\label{tj4d}}
	\renewcommand{\arraystretch}{0.7}
	\setlength{\tabcolsep}{1.5mm}{
		\begin{tabular}{cc|cc|cc|cccccc|cc|cccccc|cc}
			\hline
			\multicolumn{2}{c|}{\multirow{4}[1]{*}{Method}} & \multicolumn{2}{c|}{\multirow{4}[1]{*}{Year}} & \multicolumn{2}{c|}{\multirow{4}[1]{*}{Modality}} & \multicolumn{8}{c|}{\multirow{2}[1]{*}{3-D mAP}}               & \multicolumn{8}{c}{\multirow{2}[1]{*}{BEV mAP}} \\
			\multicolumn{2}{c|}{} & \multicolumn{2}{c|}{} & \multicolumn{2}{c|}{} & \multicolumn{8}{c|}{}                                         & \multicolumn{8}{c}{} \\
			\cline{7-22}    \multicolumn{2}{c|}{} & \multicolumn{2}{c|}{} & \multicolumn{2}{c|}{} & \multirow{2}[2]{*}{Car} & \multicolumn{2}{c}{\multirow{2}[1]{*}{Pedestrian}} & \multicolumn{2}{c}{\multirow{2}[1]{*}{Cyclist}} & \multirow{2}[1]{*}{Truck} & \multicolumn{2}{c|}{\multirow{2}[1]{*}{mAP}} & \multirow{2}[1]{*}{Car} & \multicolumn{2}{c}{\multirow{2}[1]{*}{Pedestrian}} & \multicolumn{2}{c}{\multirow{2}[1]{*}{Cyclist}} & \multirow{2}[1]{*}{Truck} & \multicolumn{2}{c}{\multirow{2}[1]{*}{mAP}} \\
			\multicolumn{2}{c|}{} & \multicolumn{2}{c|}{} & \multicolumn{2}{c|}{} &       & \multicolumn{2}{c}{} & \multicolumn{2}{c}{} &       & \multicolumn{2}{c|}{} &       & \multicolumn{2}{c}{} & \multicolumn{2}{c}{} &       & \multicolumn{2}{c}{} \\
			\hline
			\multicolumn{2}{c|}{\multirow{2}[1]{*}{PointPillars \cite{PointPillars}}} & \multicolumn{2}{c|}{\multirow{2}[1]{*}{2019}} & \multicolumn{2}{c|}{\multirow{2}[1]{*}{R}} & \multirow{2}[1]{*}{21.26 } & \multicolumn{2}{c}{\multirow{2}[1]{*}{28.33 }} & \multicolumn{2}{c}{\multirow{2}[1]{*}{52.47 }} & \multirow{2}[1]{*}{11.18 } & \multicolumn{2}{c|}{\multirow{2}[1]{*}{28.31 }} & \multirow{2}[1]{*}{38.34 } & \multicolumn{2}{c}{\multirow{2}[1]{*}{32.26 }} & \multicolumn{2}{c}{\multirow{2}[1]{*}{56.11 }} & \multirow{2}[1]{*}{18.19 } & \multicolumn{2}{c}{\multirow{2}[1]{*}{36.23 }} \\
			\multicolumn{2}{c|}{} & \multicolumn{2}{c|}{} & \multicolumn{2}{c|}{} &       & \multicolumn{2}{c}{} & \multicolumn{2}{c}{} &       & \multicolumn{2}{c|}{} &       & \multicolumn{2}{c}{} & \multicolumn{2}{c}{} &       & \multicolumn{2}{c}{} \\
			\multicolumn{2}{c|}{\multirow{2}[0]{*}{CenterPoint \cite{CenterPoint}}} & \multicolumn{2}{c|}{\multirow{2}[0]{*}{2021}} & \multicolumn{2}{c|}{\multirow{2}[0]{*}{R}} & \multirow{2}[0]{*}{22.03 } & \multicolumn{2}{c}{\multirow{2}[0]{*}{25.02 }} & \multicolumn{2}{c}{\multirow{2}[0]{*}{53.32 }} & \multirow{2}[0]{*}{15.92 } & \multicolumn{2}{c|}{\multirow{2}[0]{*}{29.07 }} & \multirow{2}[0]{*}{33.03 } & \multicolumn{2}{c}{\multirow{2}[0]{*}{27.87 }} & \multicolumn{2}{c}{\multirow{2}[0]{*}{58.74 }} & \multirow{2}[0]{*}{25.09 } & \multicolumn{2}{c}{\multirow{2}[0]{*}{36.18 }} \\
			\multicolumn{2}{c|}{} & \multicolumn{2}{c|}{} & \multicolumn{2}{c|}{} &       & \multicolumn{2}{c}{} & \multicolumn{2}{c}{} &       & \multicolumn{2}{c|}{} &       & \multicolumn{2}{c}{} & \multicolumn{2}{c}{} &       & \multicolumn{2}{c}{} \\
			\multicolumn{2}{c|}{\multirow{2}[0]{*}{RadarPillarNet \cite{RCFusion}}} & \multicolumn{2}{c|}{\multirow{2}[0]{*}{2023}} & \multicolumn{2}{c|}{\multirow{2}[0]{*}{R}} & \multirow{2}[0]{*}{28.45 } & \multicolumn{2}{c}{\multirow{2}[0]{*}{26.24 }} & \multicolumn{2}{c}{\multirow{2}[0]{*}{51.57 }} & \multirow{2}[0]{*}{15.20 } & \multicolumn{2}{c|}{\multirow{2}[0]{*}{30.37 }} & \multirow{2}[0]{*}{45.72 } & \multicolumn{2}{c}{\multirow{2}[0]{*}{29.19 }} & \multicolumn{2}{c}{\multirow{2}[0]{*}{56.89 }} & \multirow{2}[0]{*}{25.17 } & \multicolumn{2}{c}{\multirow{2}[0]{*}{39.24 }} \\
			\multicolumn{2}{c|}{} & \multicolumn{2}{c|}{} & \multicolumn{2}{c|}{} &       & \multicolumn{2}{c}{} & \multicolumn{2}{c}{} &       & \multicolumn{2}{c|}{} &       & \multicolumn{2}{c}{} & \multicolumn{2}{c}{} &       & \multicolumn{2}{c}{} \\
			\multicolumn{2}{c|}{\multirow{2}[0]{*}{SMURF \cite{SMURF}}} & \multicolumn{2}{c|}{\multirow{2}[0]{*}{2023}} & \multicolumn{2}{c|}{\multirow{2}[0]{*}{R}} & \multirow{2}[0]{*}{28.47 } & \multicolumn{2}{c}{\multirow{2}[0]{*}{26.22 }} & \multicolumn{2}{c}{\multirow{2}[0]{*}{54.61 }} & \multirow{2}[0]{*}{22.64 } & \multicolumn{2}{c|}{\multirow{2}[0]{*}{32.99 }} & \multirow{2}[0]{*}{43.13 } & \multicolumn{2}{c}{\multirow{2}[0]{*}{29.19 }} & \multicolumn{2}{c}{\multirow{2}[0]{*}{58.81 }} & \multirow{2}[0]{*}{32.80 } & \multicolumn{2}{c}{\multirow{2}[0]{*}{40.98 }} \\
			\multicolumn{2}{c|}{} & \multicolumn{2}{c|}{} & \multicolumn{2}{c|}{} &       & \multicolumn{2}{c}{} & \multicolumn{2}{c}{} &       & \multicolumn{2}{c|}{} &       & \multicolumn{2}{c}{} & \multicolumn{2}{c}{} &       & \multicolumn{2}{c}{} \\
			\multicolumn{2}{c|}{\multirow{2}[0]{*}{MUFASA \cite{MUFASA}}} & \multicolumn{2}{c|}{\multirow{2}[0]{*}{2024}} & \multicolumn{2}{c|}{\multirow{2}[0]{*}{R}} & \multirow{2}[0]{*}{27.86 } & \multicolumn{2}{c}{\multirow{2}[0]{*}{25.42 }} & \multicolumn{2}{c}{\multirow{2}[0]{*}{42.58 }} & \multirow{2}[0]{*}{19.61 } & \multicolumn{2}{c|}{\multirow{2}[0]{*}{28.87 }} & \multirow{2}[0]{*}{41.72 } & \multicolumn{2}{c}{\multirow{2}[0]{*}{27.93 }} & \multicolumn{2}{c}{\multirow{2}[0]{*}{46.42 }} & \multirow{2}[0]{*}{28.69 } & \multicolumn{2}{c}{\multirow{2}[0]{*}{36.19 }} \\
			\multicolumn{2}{c|}{} & \multicolumn{2}{c|}{} & \multicolumn{2}{c|}{} &       & \multicolumn{2}{c}{} & \multicolumn{2}{c}{} &       & \multicolumn{2}{c|}{} &       & \multicolumn{2}{c}{} & \multicolumn{2}{c}{} &       & \multicolumn{2}{c}{} \\
			\multicolumn{2}{c|}{\multirow{2}[1]{*}{MAFF-Net \cite{MAFF-Net}}} & \multicolumn{2}{c|}{\multirow{2}[1]{*}{2025}} & \multicolumn{2}{c|}{\multirow{2}[1]{*}{R}} & \multirow{2}[1]{*}{27.31 } & \multicolumn{2}{c}{\multirow{2}[1]{*}{33.13 }} & \multicolumn{2}{c}{\multirow{2}[1]{*}{54.35 }} & \multirow{2}[1]{*}{26.71 } & \multicolumn{2}{c|}{\multirow{2}[1]{*}{35.38 }} & \multirow{2}[1]{*}{39.05 } & \multicolumn{2}{c}{\multirow{2}[1]{*}{35.25 }} & \multicolumn{2}{c}{\multirow{2}[1]{*}{56.35 }} & \multirow{2}[1]{*}{35.73 } & \multicolumn{2}{c}{\multirow{2}[1]{*}{41.59 }} \\
			\multicolumn{2}{c|}{} & \multicolumn{2}{c|}{} & \multicolumn{2}{c|}{} &       & \multicolumn{2}{c}{} & \multicolumn{2}{c}{} &       & \multicolumn{2}{c|}{} &       & \multicolumn{2}{c}{} & \multicolumn{2}{c}{} &       & \multicolumn{2}{c}{} \\
			\hline
			\multicolumn{2}{c|}{\multirow{2}[1]{*}{PointAugmenting \cite{pointaugmenting}}} & \multicolumn{2}{c|}{\multirow{2}[1]{*}{2021}} & \multicolumn{2}{c|}{\multirow{2}[1]{*}{R+C}} & \multirow{2}[1]{*}{22.63 } & \multicolumn{2}{c}{\multirow{2}[1]{*}{26.23 }} & \multicolumn{2}{c}{\multirow{2}[1]{*}{53.52 }} & \multirow{2}[1]{*}{13.37 } & \multicolumn{2}{c|}{\multirow{2}[1]{*}{28.94 }} & \multirow{2}[1]{*}{43.42 } & \multicolumn{2}{c}{\multirow{2}[1]{*}{29.65 }} & \multicolumn{2}{c}{\multirow{2}[1]{*}{59.21 }} & \multirow{2}[1]{*}{23.88 } & \multicolumn{2}{c}{\multirow{2}[1]{*}{39.04 }} \\
			\multicolumn{2}{c|}{} & \multicolumn{2}{c|}{} & \multicolumn{2}{c|}{} &       & \multicolumn{2}{c}{} & \multicolumn{2}{c}{} &       & \multicolumn{2}{c|}{} &       & \multicolumn{2}{c}{} & \multicolumn{2}{c}{} &       & \multicolumn{2}{c}{} \\
			\multicolumn{2}{c|}{\multirow{2}[0]{*}{RCFusion \cite{RCFusion}}} & \multicolumn{2}{c|}{\multirow{2}[0]{*}{2023}} & \multicolumn{2}{c|}{\multirow{2}[0]{*}{R+C}} & \multirow{2}[0]{*}{29.72 } & \multicolumn{2}{c}{\multirow{2}[0]{*}{27.17 }} & \multicolumn{2}{c}{\multirow{2}[0]{*}{54.93 }} & \multirow{2}[0]{*}{23.56 } & \multicolumn{2}{c|}{\multirow{2}[0]{*}{33.85 }} & \multirow{2}[0]{*}{40.89 } & \multicolumn{2}{c}{\multirow{2}[0]{*}{30.95 }} & \multicolumn{2}{c}{\multirow{2}[0]{*}{58.30 }} & \multirow{2}[0]{*}{28.92 } & \multicolumn{2}{c}{\multirow{2}[0]{*}{39.76 }} \\
			\multicolumn{2}{c|}{} & \multicolumn{2}{c|}{} & \multicolumn{2}{c|}{} &       & \multicolumn{2}{c}{} & \multicolumn{2}{c}{} &       & \multicolumn{2}{c|}{} &       & \multicolumn{2}{c}{} & \multicolumn{2}{c}{} &       & \multicolumn{2}{c}{} \\
			\multicolumn{2}{c|}{\multirow{2}[0]{*}{FUTR3D \cite{FUTR3D}}} & \multicolumn{2}{c|}{\multirow{2}[0]{*}{2023}} & \multicolumn{2}{c|}{\multirow{2}[0]{*}{R+C}} & \multirow{2}[0]{*}{-} & \multicolumn{2}{c}{\multirow{2}[0]{*}{-}} & \multicolumn{2}{c}{\multirow{2}[0]{*}{-}} & \multirow{2}[0]{*}{-} & \multicolumn{2}{c|}{\multirow{2}[0]{*}{33.42 }} & \multirow{2}[0]{*}{-} & \multicolumn{2}{c}{\multirow{2}[0]{*}{-}} & \multicolumn{2}{c}{\multirow{2}[0]{*}{-}} & \multirow{2}[0]{*}{-} & \multicolumn{2}{c}{\multirow{2}[0]{*}{37.51 }} \\
			\multicolumn{2}{c|}{} & \multicolumn{2}{c|}{} & \multicolumn{2}{c|}{} &       & \multicolumn{2}{c}{} & \multicolumn{2}{c}{} &       & \multicolumn{2}{c|}{} &       & \multicolumn{2}{c}{} & \multicolumn{2}{c}{} &       & \multicolumn{2}{c}{} \\
			\multicolumn{2}{c|}{\multirow{2}[0]{*}{BEVFusion \cite{BEVFusion}}} & \multicolumn{2}{c|}{\multirow{2}[0]{*}{2023}} & \multicolumn{2}{c|}{\multirow{2}[0]{*}{R+C}} & \multirow{2}[0]{*}{38.09 } & \multicolumn{2}{c}{\multirow{2}[0]{*}{29.45 }} & \multicolumn{2}{c}{\multirow{2}[0]{*}{51.26 }} & \multirow{2}[0]{*}{23.73 } & \multicolumn{2}{c|}{\multirow{2}[0]{*}{35.63 }} & \multirow{2}[0]{*}{48.53 } & \multicolumn{2}{c}{\multirow{2}[0]{*}{32.04 }} & \multicolumn{2}{c}{\multirow{2}[0]{*}{55.40 }} & \multirow{2}[0]{*}{28.96 } & \multicolumn{2}{c}{\multirow{2}[0]{*}{41.23 }} \\
			\multicolumn{2}{c|}{} & \multicolumn{2}{c|}{} & \multicolumn{2}{c|}{} &       & \multicolumn{2}{c}{} & \multicolumn{2}{c}{} &       & \multicolumn{2}{c|}{} &       & \multicolumn{2}{c}{} & \multicolumn{2}{c}{} &       & \multicolumn{2}{c}{} \\
			\multicolumn{2}{c|}{\multirow{2}[0]{*}{LXL \cite{LxL}}} & \multicolumn{2}{c|}{\multirow{2}[0]{*}{2023}} & \multicolumn{2}{c|}{\multirow{2}[0]{*}{R+C}} & \multirow{2}[0]{*}{-} & \multicolumn{2}{c}{\multirow{2}[0]{*}{-}} & \multicolumn{2}{c}{\multirow{2}[0]{*}{-}} & \multirow{2}[0]{*}{-} & \multicolumn{2}{c|}{\multirow{2}[0]{*}{36.32 }} & \multirow{2}[0]{*}{-} & \multicolumn{2}{c}{\multirow{2}[0]{*}{-}} & \multicolumn{2}{c}{\multirow{2}[0]{*}{-}} & \multirow{2}[0]{*}{-} & \multicolumn{2}{c}{\multirow{2}[0]{*}{41.20 }} \\
			\multicolumn{2}{c|}{} & \multicolumn{2}{c|}{} & \multicolumn{2}{c|}{} &       & \multicolumn{2}{c}{} & \multicolumn{2}{c}{} &       & \multicolumn{2}{c|}{} &       & \multicolumn{2}{c}{} & \multicolumn{2}{c}{} &       & \multicolumn{2}{c}{} \\
			\multicolumn{2}{c|}{\multirow{2}[0]{*}{LXLv2 \cite{LxLv2}}} & \multicolumn{2}{c|}{\multirow{2}[0]{*}{2023}} & \multicolumn{2}{c|}{\multirow{2}[0]{*}{R+C}} & \multirow{2}[0]{*}{-} & \multicolumn{2}{c}{\multirow{2}[0]{*}{-}} & \multicolumn{2}{c}{\multirow{2}[0]{*}{-}} & \multirow{2}[0]{*}{-} & \multicolumn{2}{c|}{\multirow{2}[0]{*}{37.32 }} & \multirow{2}[0]{*}{-} & \multicolumn{2}{c}{\multirow{2}[0]{*}{-}} & \multicolumn{2}{c}{\multirow{2}[0]{*}{-}} & \multirow{2}[0]{*}{-} & \multicolumn{2}{c}{\multirow{2}[0]{*}{42.35 }} \\
			\multicolumn{2}{c|}{} & \multicolumn{2}{c|}{} & \multicolumn{2}{c|}{} &       & \multicolumn{2}{c}{} & \multicolumn{2}{c}{} &       & \multicolumn{2}{c|}{} &       & \multicolumn{2}{c}{} & \multicolumn{2}{c}{} &       & \multicolumn{2}{c}{} \\
			\multicolumn{2}{c|}{\multirow{2}[0]{*}{RCBEVDet \cite{RCBEVDet}}} & \multicolumn{2}{c|}{\multirow{2}[0]{*}{2024}} & \multicolumn{2}{c|}{\multirow{2}[0]{*}{R+C}} & \multirow{2}[0]{*}{29.38 } & \multicolumn{2}{c}{\multirow{2}[0]{*}{28.13 }} & \multicolumn{2}{c}{\multirow{2}[0]{*}{49.60 }} & \multirow{2}[0]{*}{17.37 } & \multicolumn{2}{c|}{\multirow{2}[0]{*}{31.12 }} & \multirow{2}[0]{*}{41.77 } & \multicolumn{2}{c}{\multirow{2}[0]{*}{28.98 }} & \multicolumn{2}{c}{\multirow{2}[0]{*}{53.03 }} & \multirow{2}[0]{*}{26.05 } & \multicolumn{2}{c}{\multirow{2}[0]{*}{37.46 }} \\
			\multicolumn{2}{c|}{} & \multicolumn{2}{c|}{} & \multicolumn{2}{c|}{} &       & \multicolumn{2}{c}{} & \multicolumn{2}{c}{} &       & \multicolumn{2}{c|}{} &       & \multicolumn{2}{c}{} & \multicolumn{2}{c}{} &       & \multicolumn{2}{c}{} \\
			\multicolumn{2}{c|}{\multirow{2}[0]{*}{RCDFNet \cite{RCDFNet}}} & \multicolumn{2}{c|}{\multirow{2}[0]{*}{2025}} & \multicolumn{2}{c|}{\multirow{2}[0]{*}{R+C}} & \multirow{2}[0]{*}{42.52 } & \multicolumn{2}{c}{\multirow{2}[0]{*}{33.77 }} & \multicolumn{2}{c}{\multirow{2}[0]{*}{55.89 }} & \multirow{2}[0]{*}{19.51 } & \multicolumn{2}{c|}{\multirow{2}[0]{*}{37.93 }} & \multirow{2}[0]{*}{54.42 } & \multicolumn{2}{c}{\multirow{2}[0]{*}{\textbf{35.81 }}} & \multicolumn{2}{c}{\multirow{2}[0]{*}{59.27 }} & \multirow{2}[0]{*}{32.93 } & \multicolumn{2}{c}{\multirow{2}[0]{*}{45.61 }} \\
			\multicolumn{2}{c|}{} & \multicolumn{2}{c|}{} & \multicolumn{2}{c|}{} &       & \multicolumn{2}{c}{} & \multicolumn{2}{c}{} &       & \multicolumn{2}{c|}{} &       & \multicolumn{2}{c}{} & \multicolumn{2}{c}{} &       & \multicolumn{2}{c}{} \\
			\multicolumn{2}{c|}{\multirow{2}[0]{*}{UniBEVFusion \cite{UniBEVFusion}}} & \multicolumn{2}{c|}{\multirow{2}[0]{*}{2025}} & \multicolumn{2}{c|}{\multirow{2}[0]{*}{R+C}} & \multirow{2}[0]{*}{44.26 } & \multicolumn{2}{c}{\multirow{2}[0]{*}{27.92 }} & \multicolumn{2}{c}{\multirow{2}[0]{*}{51.11 }} & \multirow{2}[0]{*}{27.75 } & \multicolumn{2}{c|}{\multirow{2}[0]{*}{37.76 }} & \multirow{2}[0]{*}{50.43 } & \multicolumn{2}{c}{\multirow{2}[0]{*}{29.57 }} & \multicolumn{2}{c}{\multirow{2}[0]{*}{56.48 }} & \multirow{2}[0]{*}{35.22 } & \multicolumn{2}{c}{\multirow{2}[0]{*}{42.92 }} \\
			\multicolumn{2}{c|}{} & \multicolumn{2}{c|}{} & \multicolumn{2}{c|}{} &       & \multicolumn{2}{c}{} & \multicolumn{2}{c}{} &       & \multicolumn{2}{c|}{} &       & \multicolumn{2}{c}{} & \multicolumn{2}{c}{} &       & \multicolumn{2}{c}{} \\
			\multicolumn{2}{c|}{\multirow{2}[0]{*}{HSGFusion \cite{HGSFusion}}} & \multicolumn{2}{c|}{\multirow{2}[0]{*}{2025}} & \multicolumn{2}{c|}{\multirow{2}[0]{*}{R+C}} & \multirow{2}[0]{*}{-} & \multicolumn{2}{c}{\multirow{2}[0]{*}{-}} & \multicolumn{2}{c}{\multirow{2}[0]{*}{-}} & \multirow{2}[0]{*}{-} & \multicolumn{2}{c|}{\multirow{2}[0]{*}{37.21 }} & \multirow{2}[0]{*}{-} & \multicolumn{2}{c}{\multirow{2}[0]{*}{-}} & \multicolumn{2}{c}{\multirow{2}[0]{*}{-}} & \multirow{2}[0]{*}{-} & \multicolumn{2}{c}{\multirow{2}[0]{*}{43.23 }} \\
			\multicolumn{2}{c|}{} & \multicolumn{2}{c|}{} & \multicolumn{2}{c|}{} &       & \multicolumn{2}{c}{} & \multicolumn{2}{c}{} &       & \multicolumn{2}{c|}{} &       & \multicolumn{2}{c}{} & \multicolumn{2}{c}{} &       & \multicolumn{2}{c}{} \\
			\multicolumn{2}{c|}{\multirow{2}[0]{*}{SFGFusion \cite{SFGFusion}}} & \multicolumn{2}{c|}{\multirow{2}[0]{*}{2025}} & \multicolumn{2}{c|}{\multirow{2}[0]{*}{R+C}} & \multirow{2}[0]{*}{33.05 } & \multicolumn{2}{c}{\multirow{2}[0]{*}{27.01 }} & \multicolumn{2}{c}{\multirow{2}[0]{*}{55.12 }} & \multirow{2}[0]{*}{27.10 } & \multicolumn{2}{c|}{\multirow{2}[0]{*}{35.57 }} & \multirow{2}[0]{*}{46.63 } & \multicolumn{2}{c}{\multirow{2}[0]{*}{30.04 }} & \multicolumn{2}{c}{\multirow{2}[0]{*}{59.55 }} & \multirow{2}[0]{*}{42.21 } & \multicolumn{2}{c}{\multirow{2}[0]{*}{44.66 }} \\
			\multicolumn{2}{c|}{} & \multicolumn{2}{c|}{} & \multicolumn{2}{c|}{} &       & \multicolumn{2}{c}{} & \multicolumn{2}{c}{} &       & \multicolumn{2}{c|}{} &       & \multicolumn{2}{c}{} & \multicolumn{2}{c}{} &       & \multicolumn{2}{c}{} \\
			\multicolumn{2}{c|}{\multirow{2}[0]{*}{DSFusion \cite{DSFusion}}} & \multicolumn{2}{c|}{\multirow{2}[0]{*}{2025}} & \multicolumn{2}{c|}{\multirow{2}[0]{*}{R+C}} & \multirow{2}[0]{*}{37.76 } & \multicolumn{2}{c}{\multirow{2}[0]{*}{34.09 }} & \multicolumn{2}{c}{\multirow{2}[0]{*}{49.88 }} & \multirow{2}[0]{*}{26.87 } & \multicolumn{2}{c|}{\multirow{2}[0]{*}{37.15 }} & \multirow{2}[0]{*}{49.86 } & \multicolumn{2}{c}{\multirow{2}[0]{*}{34.17 }} & \multicolumn{2}{c}{\multirow{2}[0]{*}{52.49 }} & \multirow{2}[0]{*}{32.39 } & \multicolumn{2}{c}{\multirow{2}[0]{*}{41.21 }} \\
			\multicolumn{2}{c|}{} & \multicolumn{2}{c|}{} & \multicolumn{2}{c|}{} &       & \multicolumn{2}{c}{} & \multicolumn{2}{c}{} &       & \multicolumn{2}{c|}{} &       & \multicolumn{2}{c}{} & \multicolumn{2}{c}{} &       & \multicolumn{2}{c}{} \\
			\multicolumn{2}{c|}{\multirow{2}[0]{*}{SGDet3D \cite{SGDet3D}}} & \multicolumn{2}{c|}{\multirow{2}[0]{*}{2025}} & \multicolumn{2}{c|}{\multirow{2}[0]{*}{R+C}} & \multirow{2}[0]{*}{59.43 } & \multicolumn{2}{c}{\multirow{2}[0]{*}{26.57 }} & \multicolumn{2}{c}{\multirow{2}[0]{*}{51.30 }} & \multirow{2}[0]{*}{\textbf{30.00 }} & \multicolumn{2}{c|}{\multirow{2}[0]{*}{41.82 }} & \multirow{2}[0]{*}{66.38 } & \multicolumn{2}{c}{\multirow{2}[0]{*}{29.18 }} & \multicolumn{2}{c}{\multirow{2}[0]{*}{53.72 }} & \multirow{2}[0]{*}{\textbf{39.36 }} & \multicolumn{2}{c}{\multirow{2}[0]{*}{47.16 }} \\
			\multicolumn{2}{c|}{} & \multicolumn{2}{c|}{} & \multicolumn{2}{c|}{} &       & \multicolumn{2}{c}{} & \multicolumn{2}{c}{} &       & \multicolumn{2}{c|}{} &       & \multicolumn{2}{c}{} & \multicolumn{2}{c}{} &       & \multicolumn{2}{c}{} \\
			\multicolumn{2}{c|}{\multirow{2}[0]{*}{Chen et al. \cite{DDCFusion}}} & \multicolumn{2}{c|}{\multirow{2}[0]{*}{2025}} & \multicolumn{2}{c|}{\multirow{2}[0]{*}{R+C}} & \multirow{2}[0]{*}{36.89 } & \multicolumn{2}{c}{\multirow{2}[0]{*}{28.29 }} & \multicolumn{2}{c}{\multirow{2}[0]{*}{49.64 }} & \multirow{2}[0]{*}{23.62 } & \multicolumn{2}{c|}{\multirow{2}[0]{*}{34.61 }} & \multirow{2}[0]{*}{43.73 } & \multicolumn{2}{c}{\multirow{2}[0]{*}{34.08 }} & \multicolumn{2}{c}{\multirow{2}[0]{*}{59.49 }} & \multirow{2}[0]{*}{29.70 } & \multicolumn{2}{c}{\multirow{2}[0]{*}{41.75 }} \\
			\multicolumn{2}{c|}{} & \multicolumn{2}{c|}{} & \multicolumn{2}{c|}{} &       & \multicolumn{2}{c}{} & \multicolumn{2}{c}{} &       & \multicolumn{2}{c|}{} &       & \multicolumn{2}{c}{} & \multicolumn{2}{c}{} &       & \multicolumn{2}{c}{} \\
			\multicolumn{2}{c|}{\multirow{2}[0]{*}{Ours}} & \multicolumn{2}{c|}{\multirow{2}[0]{*}{-}} & \multicolumn{2}{c|}{\multirow{2}[0]{*}{R+C}} & \multirow{2}[0]{*}{\textbf{61.24 }} & \multicolumn{2}{c}{\multirow{2}[0]{*}{\textbf{34.28 }}} & \multicolumn{2}{c}{\multirow{2}[0]{*}{\textbf{57.48 }}} & \multirow{2}[0]{*}{19.18 } & \multicolumn{2}{c|}{\multirow{2}[0]{*}{\textbf{43.05 }}} & \multirow{2}[0]{*}{\textbf{67.73 }} & \multicolumn{2}{c}{\multirow{2}[0]{*}{34.94 }} & \multicolumn{2}{c}{\multirow{2}[0]{*}{\textbf{61.32 }}} & \multirow{2}[0]{*}{33.40 } & \multicolumn{2}{c}{\multirow{2}[0]{*}{\textbf{49.33 }}} \\
			\multicolumn{2}{c|}{} & \multicolumn{2}{c|}{} & \multicolumn{2}{c|}{} &       & \multicolumn{2}{c}{} & \multicolumn{2}{c}{} &       & \multicolumn{2}{c|}{} &       & \multicolumn{2}{c}{} & \multicolumn{2}{c}{} &       & \multicolumn{2}{c}{} \\
			\hline
	\end{tabular}}
	
\end{table*}%

\begin{table*}[htbp]
	\centering
	\small
	\caption{The result of comparing with state-pf-the-art methods on Astyx HiRes 2019 test set. The best results are highlighted with black boldface. 	\label{Astyx}}
	\renewcommand{\arraystretch}{0.7}
	\setlength{\tabcolsep}{2.5mm}{
		\begin{tabular}{cc|cc|cc|cccccc|cccccc}
			\hline
			\multicolumn{2}{c|}{\multirow{4}[1]{*}{Method}} & \multicolumn{2}{c|}{\multirow{4}[1]{*}{Year}} & \multicolumn{2}{c|}{\multirow{4}[1]{*}{Modality}} & \multicolumn{6}{c|}{\multirow{2}[1]{*}{3-D mAP}} & \multicolumn{6}{c}{\multirow{2}[1]{*}{BEV mAP}} \\
			\multicolumn{2}{c|}{} & \multicolumn{2}{c|}{} & \multicolumn{2}{c|}{} & \multicolumn{6}{c|}{}                         & \multicolumn{6}{c}{} \\
			\cline{7-18}    \multicolumn{2}{c|}{} & \multicolumn{2}{c|}{} & \multicolumn{2}{c|}{} & \multicolumn{2}{c}{\multirow{2}[1]{*}{Easy}} & \multicolumn{2}{c}{\multirow{2}[1]{*}{Moderate}} & \multicolumn{2}{c|}{\multirow{2}[1]{*}{Hard}} & \multicolumn{2}{c}{\multirow{2}[1]{*}{Easy}} & \multicolumn{2}{c}{\multirow{2}[1]{*}{Moderate}} & \multicolumn{2}{c}{\multirow{2}[1]{*}{Hard}} \\
			\multicolumn{2}{c|}{} & \multicolumn{2}{c|}{} & \multicolumn{2}{c|}{} & \multicolumn{2}{c}{} & \multicolumn{2}{c}{} & \multicolumn{2}{c|}{} & \multicolumn{2}{c}{} & \multicolumn{2}{c}{} & \multicolumn{2}{c}{} \\
			\hline
			\multicolumn{2}{c|}{\multirow{2}[1]{*}{PointPillars \cite{PointPillars}}} & \multicolumn{2}{c|}{\multirow{2}[1]{*}{2019}} & \multicolumn{2}{c|}{\multirow{2}[1]{*}{R}} & \multicolumn{2}{c}{\multirow{2}[1]{*}{30.14 }} & \multicolumn{2}{c}{\multirow{2}[1]{*}{24.06 }} & \multicolumn{2}{c|}{\multirow{2}[1]{*}{21.91 }} & \multicolumn{2}{c}{\multirow{2}[1]{*}{45.66 }} & \multicolumn{2}{c}{\multirow{2}[1]{*}{36.71 }} & \multicolumn{2}{c}{\multirow{2}[1]{*}{35.30 }} \\
			\multicolumn{2}{c|}{} & \multicolumn{2}{c|}{} & \multicolumn{2}{c|}{} & \multicolumn{2}{c}{} & \multicolumn{2}{c}{} & \multicolumn{2}{c|}{} & \multicolumn{2}{c}{} & \multicolumn{2}{c}{} & \multicolumn{2}{c}{} \\
			\multicolumn{2}{c|}{\multirow{2}[0]{*}{PointRCNN \cite{PointRCNN}}} & \multicolumn{2}{c|}{\multirow{2}[0]{*}{2019}} & \multicolumn{2}{c|}{\multirow{2}[0]{*}{R}} & \multicolumn{2}{c}{\multirow{2}[0]{*}{12.23 }} & \multicolumn{2}{c}{\multirow{2}[0]{*}{9.10 }} & \multicolumn{2}{c|}{\multirow{2}[0]{*}{9.10 }} & \multicolumn{2}{c}{\multirow{2}[0]{*}{14.95 }} & \multicolumn{2}{c}{\multirow{2}[0]{*}{13.82 }} & \multicolumn{2}{c}{\multirow{2}[0]{*}{13.89 }} \\
			\multicolumn{2}{c|}{} & \multicolumn{2}{c|}{} & \multicolumn{2}{c|}{} & \multicolumn{2}{c}{} & \multicolumn{2}{c}{} & \multicolumn{2}{c|}{} & \multicolumn{2}{c}{} & \multicolumn{2}{c}{} & \multicolumn{2}{c}{} \\
			\multicolumn{2}{c|}{\multirow{2}[0]{*}{RPFANet \cite{RPFANet}}} & \multicolumn{2}{c|}{\multirow{2}[0]{*}{2021}} & \multicolumn{2}{c|}{\multirow{2}[0]{*}{R}} & \multicolumn{2}{c}{\multirow{2}[0]{*}{38.85 }} & \multicolumn{2}{c}{\multirow{2}[0]{*}{32.19 }} & \multicolumn{2}{c|}{\multirow{2}[0]{*}{30.57 }} & \multicolumn{2}{c}{\multirow{2}[0]{*}{50.42 }} & \multicolumn{2}{c}{\multirow{2}[0]{*}{42.23 }} & \multicolumn{2}{c}{\multirow{2}[0]{*}{40.96 }} \\
			\multicolumn{2}{c|}{} & \multicolumn{2}{c|}{} & \multicolumn{2}{c|}{} & \multicolumn{2}{c}{} & \multicolumn{2}{c}{} & \multicolumn{2}{c|}{} & \multicolumn{2}{c}{} & \multicolumn{2}{c}{} & \multicolumn{2}{c}{} \\
			\multicolumn{2}{c|}{\multirow{2}[0]{*}{SECOND \cite{SECOND}}} & \multicolumn{2}{c|}{\multirow{2}[0]{*}{2018}} & \multicolumn{2}{c|}{\multirow{2}[0]{*}{R}} & \multicolumn{2}{c}{\multirow{2}[0]{*}{24.11 }} & \multicolumn{2}{c}{\multirow{2}[0]{*}{18.50 }} & \multicolumn{2}{c|}{\multirow{2}[0]{*}{17.77 }} & \multicolumn{2}{c}{\multirow{2}[0]{*}{41.25 }} & \multicolumn{2}{c}{\multirow{2}[0]{*}{30.58 }} & \multicolumn{2}{c}{\multirow{2}[0]{*}{29.33 }} \\
			\multicolumn{2}{c|}{} & \multicolumn{2}{c|}{} & \multicolumn{2}{c|}{} & \multicolumn{2}{c}{} & \multicolumn{2}{c}{} & \multicolumn{2}{c|}{} & \multicolumn{2}{c}{} & \multicolumn{2}{c}{} & \multicolumn{2}{c}{} \\
			\multicolumn{2}{c|}{\multirow{2}[0]{*}{PVRCNN \cite{PVRCNN}}} & \multicolumn{2}{c|}{\multirow{2}[0]{*}{2020}} & \multicolumn{2}{c|}{\multirow{2}[0]{*}{R}} & \multicolumn{2}{c}{\multirow{2}[0]{*}{28.21 }} & \multicolumn{2}{c}{\multirow{2}[0]{*}{22.29 }} & \multicolumn{2}{c|}{\multirow{2}[0]{*}{20.40 }} & \multicolumn{2}{c}{\multirow{2}[0]{*}{46.62 }} & \multicolumn{2}{c}{\multirow{2}[0]{*}{35.10 }} & \multicolumn{2}{c}{\multirow{2}[0]{*}{33.67 }} \\
			\multicolumn{2}{c|}{} & \multicolumn{2}{c|}{} & \multicolumn{2}{c|}{} & \multicolumn{2}{c}{} & \multicolumn{2}{c}{} & \multicolumn{2}{c|}{} & \multicolumn{2}{c}{} & \multicolumn{2}{c}{} & \multicolumn{2}{c}{} \\
			\multicolumn{2}{c|}{\multirow{2}[0]{*}{Duan et al. \cite{duan}}} & \multicolumn{2}{c|}{\multirow{2}[0]{*}{2025}} & \multicolumn{2}{c|}{\multirow{2}[0]{*}{R}} & \multicolumn{2}{c}{\multirow{2}[0]{*}{38.99 }} & \multicolumn{2}{c}{\multirow{2}[0]{*}{33.63 }} & \multicolumn{2}{c|}{\multirow{2}[0]{*}{31.89 }} & \multicolumn{2}{c}{\multirow{2}[0]{*}{51.99 }} & \multicolumn{2}{c}{\multirow{2}[0]{*}{43.93 }} & \multicolumn{2}{c}{\multirow{2}[0]{*}{42.87 }} \\
			\multicolumn{2}{c|}{} & \multicolumn{2}{c|}{} & \multicolumn{2}{c|}{} & \multicolumn{2}{c}{} & \multicolumn{2}{c}{} & \multicolumn{2}{c|}{} & \multicolumn{2}{c}{} & \multicolumn{2}{c}{} & \multicolumn{2}{c}{} \\
			\multicolumn{2}{c|}{\multirow{2}[0]{*}{Voxel-Mamba \cite{VoxelMamba}}} & \multicolumn{2}{c|}{\multirow{2}[0]{*}{2025}} & \multicolumn{2}{c|}{\multirow{2}[0]{*}{R}} & \multicolumn{2}{c}{\multirow{2}[0]{*}{20.63	}} & \multicolumn{2}{c}{\multirow{2}[0]{*}{14.70 }} & \multicolumn{2}{c|}{\multirow{2}[0]{*}{13.89	}} & \multicolumn{2}{c}{\multirow{2}[0]{*}{36.27	}} & \multicolumn{2}{c}{\multirow{2}[0]{*}{27.66	}} & \multicolumn{2}{c}{\multirow{2}[0]{*}{25.87 }} \\
			\multicolumn{2}{c|}{} & \multicolumn{2}{c|}{} & \multicolumn{2}{c|}{} & \multicolumn{2}{c}{} & \multicolumn{2}{c}{} & \multicolumn{2}{c|}{} & \multicolumn{2}{c}{} & \multicolumn{2}{c}{} & \multicolumn{2}{c}{} \\
			\hline
			\multicolumn{2}{c|}{\multirow{2}[1]{*}{SGDet3D \cite{SGDet3D}}} & \multicolumn{2}{c|}{\multirow{2}[1]{*}{2025}} & \multicolumn{2}{c|}{\multirow{2}[1]{*}{C+R}} & \multicolumn{2}{c}{\multirow{2}[1]{*}{40.62 }} & \multicolumn{2}{c}{\multirow{2}[1]{*}{36.25 }} & \multicolumn{2}{c|}{\multirow{2}[1]{*}{29.25 }} & \multicolumn{2}{c}{\multirow{2}[1]{*}{49.98 }} & \multicolumn{2}{c}{\multirow{2}[1]{*}{48.07 }} & \multicolumn{2}{c}{\multirow{2}[1]{*}{43.76 }} \\
			\multicolumn{2}{c|}{} & \multicolumn{2}{c|}{} & \multicolumn{2}{c|}{} & \multicolumn{2}{c}{} & \multicolumn{2}{c}{} & \multicolumn{2}{c|}{} & \multicolumn{2}{c}{} & \multicolumn{2}{c}{} & \multicolumn{2}{c}{} \\
			\multicolumn{2}{c|}{\multirow{2}[0]{*}{HGSFusion \cite{HGSFusion}}} & \multicolumn{2}{c|}{\multirow{2}[0]{*}{2025}} & \multicolumn{2}{c|}{\multirow{2}[0]{*}{C+R}} & \multicolumn{2}{c}{\multirow{2}[0]{*}{45.16 }} & \multicolumn{2}{c}{\multirow{2}[0]{*}{38.65 }} & \multicolumn{2}{c|}{\multirow{2}[0]{*}{32.73 }} & \multicolumn{2}{c}{\multirow{2}[0]{*}{55.81 }} & \multicolumn{2}{c}{\multirow{2}[0]{*}{47.60 }} & \multicolumn{2}{c}{\multirow{2}[0]{*}{45.18 }} \\
			\multicolumn{2}{c|}{} & \multicolumn{2}{c|}{} & \multicolumn{2}{c|}{} & \multicolumn{2}{c}{} & \multicolumn{2}{c}{} & \multicolumn{2}{c|}{} & \multicolumn{2}{c}{} & \multicolumn{2}{c}{} & \multicolumn{2}{c}{} \\
			\multicolumn{2}{c|}{\multirow{2}[0]{*}{Ours}} & \multicolumn{2}{c|}{\multirow{2}[0]{*}{-}} & \multicolumn{2}{c|}{\multirow{2}[0]{*}{C+R}} & \multicolumn{2}{c}{\multirow{2}[0]{*}{\textbf{49.61 }}} & \multicolumn{2}{c}{\multirow{2}[0]{*}{\textbf{39.28 }}} & \multicolumn{2}{c|}{\multirow{2}[0]{*}{\textbf{35.40 }}} & \multicolumn{2}{c}{\multirow{2}[0]{*}{\textbf{57.08 }}} & \multicolumn{2}{c}{\multirow{2}[0]{*}{\textbf{49.35 }}} & \multicolumn{2}{c}{\multirow{2}[0]{*}{\textbf{45.27 }}} \\
			\multicolumn{2}{c|}{} & \multicolumn{2}{c|}{} & \multicolumn{2}{c|}{} & \multicolumn{2}{c}{} & \multicolumn{2}{c}{} & \multicolumn{2}{c|}{} & \multicolumn{2}{c}{} & \multicolumn{2}{c}{} & \multicolumn{2}{c}{} \\
			\hline
	\end{tabular}}
	
\end{table*}%

\begin{figure*}[t]
	\centering
	\includegraphics[width=7in]{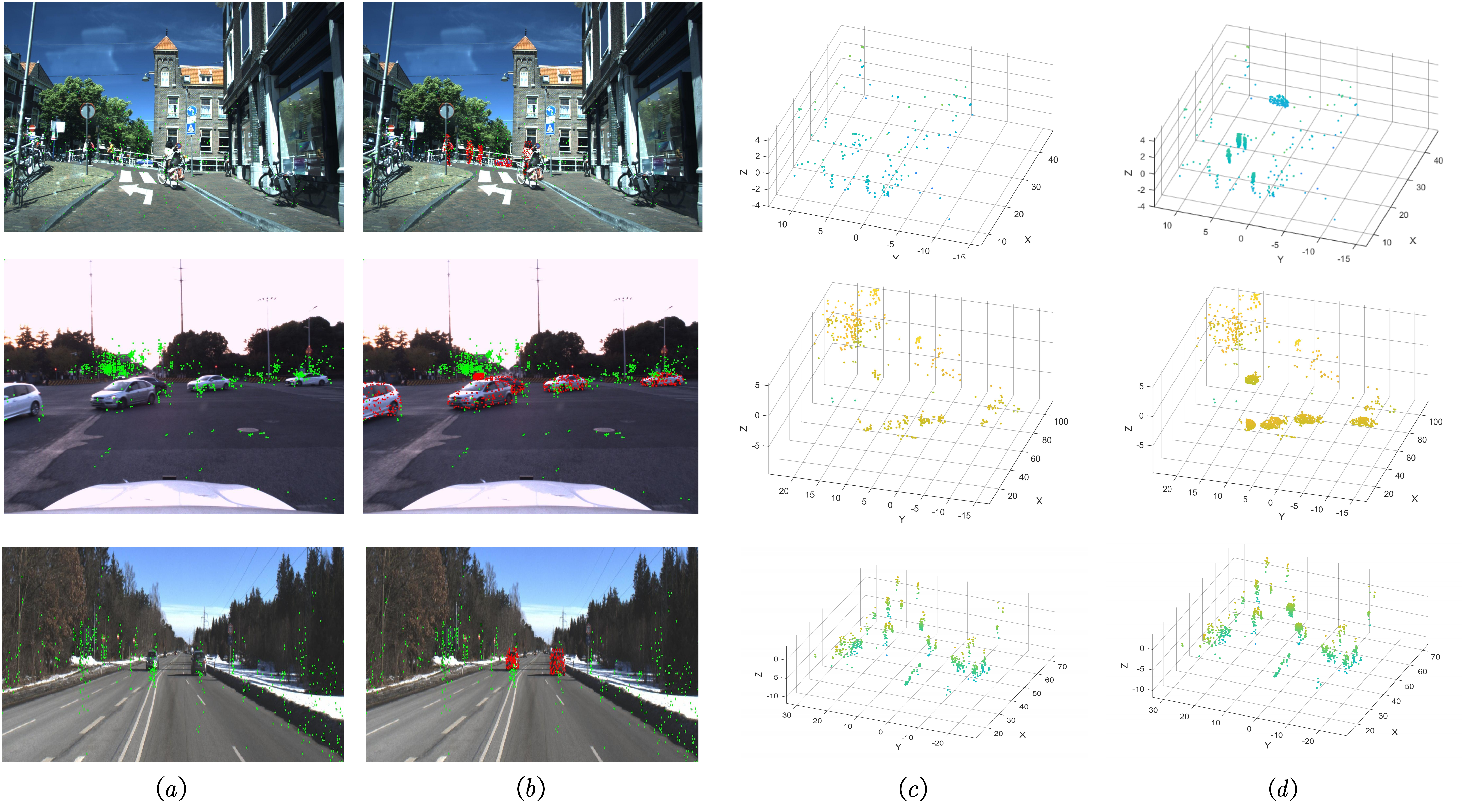}
	\caption{The visualization of radar point clouds generated by SimDen module. 
		Every dataset has a scene with one row. From top to bottom, they are VoD, TJ4DRadSet and Astyx HiRes 2019 dataset.
		(a), (b) is the projected radar point clouds on the RGB Images, the green points is the original point clouds, while the red points is generated by SimDen module. (c), (d) are the 3-D point clouds before and after SimDen's generation. \label{pc1}
	}
\end{figure*}
\begin{table}[t]
	\centering
	\small
	\caption{The Comparison of computational complexity.	\label{complex}}
	\renewcommand{\arraystretch}{0.8}
	\setlength{\tabcolsep}{0.8mm}{
		\begin{tabular}{cc|cc|cc|cc|cc|cc}
			\hline
			\multicolumn{2}{c|}{\multirow{2}[1]{*}{Model}} & \multicolumn{2}{c|}{\multirow{2}[1]{*}{Year}} & \multicolumn{2}{c|}{\multirow{2}[1]{*}{Params(MB)}} & \multicolumn{2}{c|}{\multirow{2}[1]{*}{FPS}} & \multicolumn{2}{c|}{\multirow{2}[1]{*}{EAA mAP}} & \multicolumn{2}{c}{\multirow{2}[1]{*}{DC mAP}} \\
			\multicolumn{2}{c|}{} & \multicolumn{2}{c|}{} & \multicolumn{2}{c|}{} & \multicolumn{2}{c|}{} & \multicolumn{2}{c|}{} & \multicolumn{2}{c}{} \\
			\hline
			\multicolumn{2}{c|}{\multirow{2}[1]{*}{SGDet3D \cite{SGDet3D}}} & \multicolumn{2}{c|}{\multirow{2}[1]{*}{2025}} & \multicolumn{2}{c|}{\multirow{2}[1]{*}{73.65}} & \multicolumn{2}{c|}{\multirow{2}[1]{*}{9.2}} & \multicolumn{2}{c|}{\multirow{2}[1]{*}{59.75}} & \multicolumn{2}{c}{\multirow{2}[1]{*}{77.42}} \\
			\multicolumn{2}{c|}{} & \multicolumn{2}{c|}{} & \multicolumn{2}{c|}{} & \multicolumn{2}{c|}{} & \multicolumn{2}{c|}{} & \multicolumn{2}{c}{} \\
			\multicolumn{2}{c|}{\multirow{2}[0]{*}{HGSFusion \cite{HGSFusion}}} & \multicolumn{2}{c|}{\multirow{2}[0]{*}{2025}} & \multicolumn{2}{c|}{\multirow{2}[0]{*}{64.58}} & \multicolumn{2}{c|}{\multirow{2}[0]{*}{10.3}} & \multicolumn{2}{c|}{\multirow{2}[0]{*}{57.72}} & \multicolumn{2}{c}{\multirow{2}[0]{*}{77.79}} \\
			\multicolumn{2}{c|}{} & \multicolumn{2}{c|}{} & \multicolumn{2}{c|}{} & \multicolumn{2}{c|}{} & \multicolumn{2}{c|}{} & \multicolumn{2}{c}{} \\
			\multicolumn{2}{c|}{\multirow{2}[0]{*}{IS-Fusion \cite{IS-Fusion}}} & \multicolumn{2}{c|}{\multirow{2}[0]{*}{2024}} & \multicolumn{2}{c|}{\multirow{2}[0]{*}{48.37}} & \multicolumn{2}{c|}{\multirow{2}[0]{*}{7.2}} & \multicolumn{2}{c|}{\multirow{2}[0]{*}{54.40}} & \multicolumn{2}{c}{\multirow{2}[0]{*}{74.75}} \\
			\multicolumn{2}{c|}{} & \multicolumn{2}{c|}{} & \multicolumn{2}{c|}{} & \multicolumn{2}{c|}{} & \multicolumn{2}{c|}{} & \multicolumn{2}{c}{} \\
			\multicolumn{2}{c|}{\multirow{2}[0]{*}{SFGFusion \cite{SFGFusion}}} & \multicolumn{2}{c|}{\multirow{2}[0]{*}{2025}} & \multicolumn{2}{c|}{\multirow{2}[0]{*}{N/A}} & \multicolumn{2}{c|}{\multirow{2}[0]{*}{6.4}} & \multicolumn{2}{c|}{\multirow{2}[0]{*}{55.76}} & \multicolumn{2}{c}{\multirow{2}[0]{*}{73.77}} \\
			\multicolumn{2}{c|}{} & \multicolumn{2}{c|}{} & \multicolumn{2}{c|}{} & \multicolumn{2}{c|}{} & \multicolumn{2}{c|}{} & \multicolumn{2}{c}{} \\
			\multicolumn{2}{c|}{\multirow{2}[0]{*}{RCDFNet \cite{RCDFNet}}} & \multicolumn{2}{c|}{\multirow{2}[0]{*}{2025}} & \multicolumn{2}{c|}{\multirow{2}[0]{*}{N/A}} & \multicolumn{2}{c|}{\multirow{2}[0]{*}{12.9}} & \multicolumn{2}{c|}{\multirow{2}[0]{*}{56.66}} & \multicolumn{2}{c}{\multirow{2}[0]{*}{70.61}} \\
			\multicolumn{2}{c|}{} & \multicolumn{2}{c|}{} & \multicolumn{2}{c|}{} & \multicolumn{2}{c|}{} & \multicolumn{2}{c|}{} & \multicolumn{2}{c}{} \\
			\multicolumn{2}{c|}{\multirow{2}[0]{*}{ZFusion \cite{ZFusion}}} & \multicolumn{2}{c|}{\multirow{2}[0]{*}{2025}} & \multicolumn{2}{c|}{\multirow{2}[0]{*}{N/A}} & \multicolumn{2}{c|}{\multirow{2}[0]{*}{8.8}} & \multicolumn{2}{c|}{\multirow{2}[0]{*}{51.14}} & \multicolumn{2}{c}{\multirow{2}[0]{*}{74.38}} \\
			\multicolumn{2}{c|}{} & \multicolumn{2}{c|}{} & \multicolumn{2}{c|}{} & \multicolumn{2}{c|}{} & \multicolumn{2}{c|}{} & \multicolumn{2}{c}{} \\
			\multicolumn{2}{c|}{\multirow{2}[0]{*}{Ours}} & \multicolumn{2}{c|}{\multirow{2}[0]{*}{-}} & \multicolumn{2}{c|}{\multirow{2}[0]{*}{\textbf{39.11}}} & \multicolumn{2}{c|}{\multirow{2}[0]{*}{\textbf{14.2}}} & \multicolumn{2}{c|}{\multirow{2}[0]{*}{\textbf{60.02}}} & \multicolumn{2}{c}{\multirow{2}[0]{*}{\textbf{81.78}}} \\
			\multicolumn{2}{c|}{} & \multicolumn{2}{c|}{} & \multicolumn{2}{c|}{} & \multicolumn{2}{c|}{} & \multicolumn{2}{c|}{} & \multicolumn{2}{c}{} \\
			\hline
		\end{tabular}
	}
	
\end{table}%

\subsubsection{The Comparison of Computational Complexity}
Table \ref{complex} shows the comparison of computational complexity. The models chosen for comparison are mainly proposed in 2024-2025. The chosen metrics are parameter quantity in units of MegaByte (MB), as well as Frame-Per-Second (FPS) during inference.
Notably, all of the comparisons are under FP16.
It could be seen that our model contains 39.11 MB parameters, which is lower about 40.54, 25.47, 9.26 MB than SGDet3D, HGSFusion and IS-Fusion \cite{IS-Fusion} respectively.
Moreover, our method also outperforms in terms of FPS, with about 1.3, 3.9 and 5 imrpovement when comparing with RCDFNet \cite{RCDFNet}, HGSFusion and SGDet3D. 
At the same time, the detection precisions of our method on the VoD dataset are the best in these compared models.

This experiment shows that our model owns lower parameter quantity and faster inference speed. This is mainly because our method avoids using mechanisms with high computational complexity such as Transformer  \cite{SGDet3D, ZFusion} and Cross-Attention \cite{RCFusion, RCDFNet} for feature extraction and multi-modality fusion.

\subsection{Ablation Study}
In this section, the results of ablation study are introduced, which is conducted on the VoD dataset. The chosen metrics are EAA mAP and DC mAP.

\subsubsection{Ablation Study on Proposed Modules}
Table \ref{als_module} shows the ablation study results on proposed modules. Through the results of Method 0, Method 1, Method 2 and Method 3, it could be seen that our proposed modules bring gains for EAA mAP and DC mAP. Notably, for SimDen, it brings -1.84 improvements in EAA mAP, mainly because the depth filtering depends on the mode value of point clouds depth. When the instances are out of DC, the distribution of depth values are influnced by the background behind the instances more easily. Nonetheless, SimDen module brings 1.15 improvements in DC mAP.
For RCM and MMIF module, they bring certain gains for two metrics obviously.

For Method 4, comparing with Method 1, it could be seen that RCM could bring about 2.35 points imrpovements for EAA mAP, while the model's Performance declines without MMIF module, comparing with Method 7. These results shows the importance of reducing the heterogeneity among modalities, which is achieved by MMIF module. 
This effect  also could be seen when comparing Method 1 and Method 5. 

The combined effect of RCM and MMIF module could be seen. Comparing with Method 2 and Method 6, only achieving feature compensation could bring less improvement because the heterogeneity among modalities cause interference. Comparing with Method 1, Method 3 and Method 6, lacking enough point clouds information is hard for multi-modality fusion, though there are small gains when only adopting proposed modules respectively.

The method 7 achieves the best performance by combining all of the proposed modules, which shows that the dense radar point clouds, unified samantic space and reducing modalities' heterogeneity are improtant for 3-D object detection.

\begin{table}[t]
	\centering
	\small
	\caption{Ablation study on the proposed modules in SDCM. 	\label{als_module}}
	\renewcommand{\arraystretch}{0.7}
	\setlength{\tabcolsep}{1.5mm}{
		\begin{tabular}{cc|cccccc|cc|cc}
			\hline
			\multicolumn{2}{c|}{\multirow{4}[1]{*}{Method}} & \multicolumn{2}{c}{\multirow{4}[1]{*}{SimDen }} & \multicolumn{2}{c}{\multirow{4}[1]{*}{ RCM}} & \multicolumn{2}{c|}{\multirow{4}[1]{*}{MMIF}} & \multicolumn{2}{c|}{\multirow{4}[1]{*}{EAA mAP}} & \multicolumn{2}{c}{\multirow{4}[1]{*}{DC mAP}} \\
			\multicolumn{2}{c|}{} & \multicolumn{2}{c}{} & \multicolumn{2}{c}{} & \multicolumn{2}{c|}{} & \multicolumn{2}{c|}{} & \multicolumn{2}{c}{} \\
			\multicolumn{2}{c|}{} & \multicolumn{2}{c}{} & \multicolumn{2}{c}{} & \multicolumn{2}{c|}{} & \multicolumn{2}{c|}{} & \multicolumn{2}{c}{} \\
			\multicolumn{2}{c|}{} & \multicolumn{2}{c}{} & \multicolumn{2}{c}{} & \multicolumn{2}{c|}{} & \multicolumn{2}{c|}{} & \multicolumn{2}{c}{} \\
			\hline
			\multicolumn{2}{c|}{\multirow{2}[1]{*}{Method 0}} & \multicolumn{2}{c}{\multirow{2}[1]{*}{}} & \multicolumn{2}{c}{\multirow{2}[1]{*}{}} & \multicolumn{2}{c|}{\multirow{2}[1]{*}{}} & \multicolumn{2}{c|}{\multirow{2}[1]{*}{57.72}} & \multicolumn{2}{c}{\multirow{2}[1]{*}{77.79}} \\
			\multicolumn{2}{c|}{} & \multicolumn{2}{c}{} & \multicolumn{2}{c}{} & \multicolumn{2}{c|}{} & \multicolumn{2}{c|}{} & \multicolumn{2}{c}{} \\
			\multicolumn{2}{c|}{\multirow{2}[0]{*}{Method 1}} & \multicolumn{2}{c}{\multirow{2}[0]{*}{\checkmark}} & \multicolumn{2}{c}{\multirow{2}[0]{*}{}} & \multicolumn{2}{c|}{\multirow{2}[0]{*}{}} & \multicolumn{2}{c|}{\multirow{2}[0]{*}{55.88}} & \multicolumn{2}{c}{\multirow{2}[0]{*}{78.94}} \\
			\multicolumn{2}{c|}{} & \multicolumn{2}{c}{} & \multicolumn{2}{c}{} & \multicolumn{2}{c|}{} & \multicolumn{2}{c|}{} & \multicolumn{2}{c}{} \\
			\multicolumn{2}{c|}{\multirow{2}[0]{*}{Method 2}} & \multicolumn{2}{c}{\multirow{2}[0]{*}{}} & \multicolumn{2}{c}{\multirow{2}[0]{*}{\checkmark}} & \multicolumn{2}{c|}{\multirow{2}[0]{*}{}} & \multicolumn{2}{c|}{\multirow{2}[0]{*}{57.95}} & \multicolumn{2}{c}{\multirow{2}[0]{*}{78.33}} \\
			\multicolumn{2}{c|}{} & \multicolumn{2}{c}{} & \multicolumn{2}{c}{} & \multicolumn{2}{c|}{} & \multicolumn{2}{c|}{} & \multicolumn{2}{c}{} \\
			\multicolumn{2}{c|}{\multirow{2}[0]{*}{Method 3}} & \multicolumn{2}{c}{\multirow{2}[0]{*}{}} & \multicolumn{2}{c}{\multirow{2}[0]{*}{}} & \multicolumn{2}{c|}{\multirow{2}[0]{*}{\checkmark}} & \multicolumn{2}{c|}{\multirow{2}[0]{*}{58.44}} & \multicolumn{2}{c}{\multirow{2}[0]{*}{78.28}} \\
			\multicolumn{2}{c|}{} & \multicolumn{2}{c}{} & \multicolumn{2}{c}{} & \multicolumn{2}{c|}{} & \multicolumn{2}{c|}{} & \multicolumn{2}{c}{} \\
			\multicolumn{2}{c|}{\multirow{2}[0]{*}{Method 4}} & \multicolumn{2}{c}{\multirow{2}[0]{*}{\checkmark}} & \multicolumn{2}{c}{\multirow{2}[0]{*}{\checkmark}} & \multicolumn{2}{c|}{\multirow{2}[0]{*}{}} & \multicolumn{2}{c|}{\multirow{2}[0]{*}{58.23}} & \multicolumn{2}{c}{\multirow{2}[0]{*}{78.45}} \\
			\multicolumn{2}{c|}{} & \multicolumn{2}{c}{} & \multicolumn{2}{c}{} & \multicolumn{2}{c|}{} & \multicolumn{2}{c|}{} & \multicolumn{2}{c}{} \\
			\multicolumn{2}{c|}{\multirow{2}[0]{*}{Method 5}} & \multicolumn{2}{c}{\multirow{2}[0]{*}{\checkmark}} & \multicolumn{2}{c}{\multirow{2}[0]{*}{}} & \multicolumn{2}{c|}{\multirow{2}[0]{*}{\checkmark}} & \multicolumn{2}{c|}{\multirow{2}[0]{*}{58.18}} & \multicolumn{2}{c}{\multirow{2}[0]{*}{78.72}} \\
			\multicolumn{2}{c|}{} & \multicolumn{2}{c}{} & \multicolumn{2}{c}{} & \multicolumn{2}{c|}{} & \multicolumn{2}{c|}{} & \multicolumn{2}{c}{} \\
			\multicolumn{2}{c|}{\multirow{2}[0]{*}{Method 6}} & \multicolumn{2}{c}{\multirow{2}[0]{*}{}} & \multicolumn{2}{c}{\multirow{2}[0]{*}{\checkmark}} & \multicolumn{2}{c|}{\multirow{2}[0]{*}{\checkmark}} & \multicolumn{2}{c|}{\multirow{2}[0]{*}{58.34}} & \multicolumn{2}{c}{\multirow{2}[0]{*}{77.56}} \\
			\multicolumn{2}{c|}{} & \multicolumn{2}{c}{} & \multicolumn{2}{c}{} & \multicolumn{2}{c|}{} & \multicolumn{2}{c|}{} & \multicolumn{2}{c}{} \\
			\multicolumn{2}{c|}{\multirow{2}[0]{*}{Method 7}} & \multicolumn{2}{c}{\multirow{2}[0]{*}{\checkmark}} & \multicolumn{2}{c}{\multirow{2}[0]{*}{\checkmark}} & \multicolumn{2}{c|}{\multirow{2}[0]{*}{\checkmark}} & \multicolumn{2}{c|}{\multirow{2}[0]{*}{\textbf{60.02}}} & \multicolumn{2}{c}{\multirow{2}[0]{*}{\textbf{81.78}}} \\
			\multicolumn{2}{c|}{} & \multicolumn{2}{c}{} & \multicolumn{2}{c}{} & \multicolumn{2}{c|}{} & \multicolumn{2}{c|}{} & \multicolumn{2}{c}{} \\
			\hline
	\end{tabular}}
	
\end{table}%
\subsubsection{Ablation study of SimDen module}
Table \ref{als_simden} shows the ablation study on the hyperparameters of SimDen module.
The main parameters are bandwidth rule and kernel function. The setting of bandwidth affects the degree of noise introduction, while the choice of kernel function does not have an obvious impact on the estimated value theoretically.

The experiment results show that, when adopting Silverman \cite{silverman} rule, the detection mAP reaches the best. However, comparing with adopting Scott \cite{scott} rule, the detection mAP is approximate. Therefore, the average bandwidth generated by these two rules has small effect on the overall detection performance. The User-Defined groups are conducted for comparison. When defining a consistent bandwidth like 0.5 and 1.0, the detection mAPs start to decrease, because the number of point clouds are random in every frame, which requires variable bandwidths for estimation, or the biases could accmulate and bring more noise points. According to the results, Silverman rule is adopted for densifying task. 

The affect of kernel function is shown in this experiment. When adopting Epanechnikov kernel, the metrics is lower than adopting Gauss kernel with about 1.29 points and 1.10 points. For Uniform kernel, the DC mAP is approximate to Gauss kernel, with the bias about 0.47 points. The Triangle kernel has the highest biases when comparing with Gauss kernel, which might be because the property of low smooth. Following the experiment results, the Gauss kernel is adopted for 3-D KDE program in the densifying task.

The visualizations of radar point clouds are shown in Fig \ref{pc1}. Every dataset has a scene with one row. From top to bottom, they are VoD, TJ4DRadSet and Astyx HiRes 2019 dataset. The green points denote the original point clouds, while the red points denote point clouds generated by SimDen module. Comparing with (a) and (b), it is obvious that the instances has more point clouds, which contains the measurement information about real time instances. The radar point clouds also could be seen under 3-D perspective. Comparing with (c) and (d), the instances have hardly point clouds information, through SimDen module, the lost information is supplied. 

\begin{figure*}[htbp]
	\centering
	\includegraphics[width=7in]{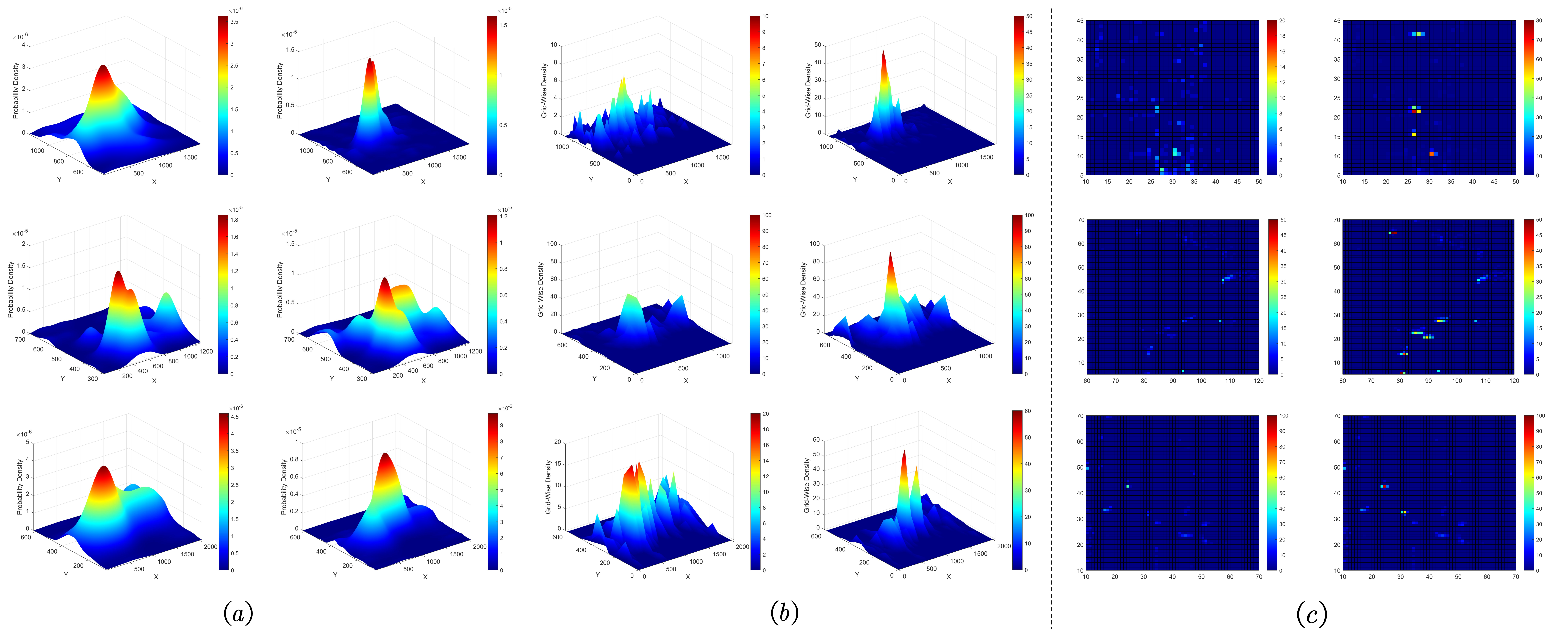}
	\caption{The 3-D probability density surface, 2-D grid-wise density surface and BEV point clouds' quantity statistics of the scenes in Fig 2. Every dataset has a scene with one row. From top to bottom, they are VoD, TJ4DRadSet and Astyx HiRes 2019 dataset. In (a), the left and right sub-figures denote the 3-D probability density surfaces of Fig.2 (c) and Fig.2 (d) respectively. In (b), the left and right sub-figures denote the 2-D grid-wise density surfaces of Fig.2 (a) and Fig.2 (b) respectively. In (c), the left and right sub-figures denote BEV point clouds' quantity statistics of Fig.2 (a) and Fig.2 (b) respectively, the size of pillars in (c) is 1 m $\times$ 1 m.  \label{surface}	}
\end{figure*}

From the perspective of 3-D probability density and 2-D grid-wise density surfaces could gain a deeper understanding of densifying's advantage, as shown in Fig \ref{surface}. 
Notably, the 2-D grid-wise density surfaces are obtained through dividing RGB images with uniform grid, and mapping the number of 2-D points within the grid onto colormap. In (a), the left sub-figures show that the gradient of original 3-D probability density surfaces decrease slowly from dense centers which may represent the location of objects.
Therefore, the background within vicinity may bring interference for feature learning. Instead, the right sub-figures show that the dense centers are obvious with rapidly decreasing surface gradient. In addition, original local dense centers are suppressed, such as the second row. Therefore, the features of the objects are enhanced. In (b), there are many local peaks on the surfaces, which supressed the information of objects. After densifying, the number of 2-D points on the objects increases and the peaks of background points are supressed.

\begin{table}[t]
	\centering
	\small
	\caption{Ablation study on the hyperparameters of SimDen module.	\label{als_simden}}
	\renewcommand{\arraystretch}{0.7}
	\setlength{\tabcolsep}{3.5mm}{
		\begin{tabular}{cc|cc|cc}
			\hline
			\multicolumn{2}{c|}{\multirow{2}[1]{*}{Bandwidth Rule}} & \multicolumn{2}{c|}{\multirow{2}[1]{*}{EAA mAP}} & \multicolumn{2}{c}{\multirow{2}[1]{*}{DC mAP}} \\
			\multicolumn{2}{c|}{} & \multicolumn{2}{c|}{} & \multicolumn{2}{c}{} \\
			\hline
			\multicolumn{2}{c|}{\multirow{2}[1]{*}{Scott \cite{scott}}} & \multicolumn{2}{c|}{\multirow{2}[1]{*}{59.66}} & \multicolumn{2}{c}{\multirow{2}[1]{*}{81.11}} \\
			\multicolumn{2}{c|}{} & \multicolumn{2}{c|}{} & \multicolumn{2}{c}{} \\
			\multicolumn{2}{c|}{\multirow{2}[0]{*}{User-Defined: 1.0}} & \multicolumn{2}{c|}{\multirow{2}[0]{*}{57.35}} & \multicolumn{2}{c}{\multirow{2}[0]{*}{79.80}} \\
			\multicolumn{2}{c|}{} & \multicolumn{2}{c|}{} & \multicolumn{2}{c}{} \\
			\multicolumn{2}{c|}{\multirow{2}[0]{*}{User-Defined: 0.5}} & \multicolumn{2}{c|}{\multirow{2}[0]{*}{58.84}} & \multicolumn{2}{c}{\multirow{2}[0]{*}{80.93}} \\
			\multicolumn{2}{c|}{} & \multicolumn{2}{c|}{} & \multicolumn{2}{c}{} \\
			\multicolumn{2}{c|}{\multirow{2}[0]{*}{Silverman \cite{silverman}}} & \multicolumn{2}{c|}{\multirow{2}[0]{*}{\textbf{60.02}}} & \multicolumn{2}{c}{\multirow{2}[0]{*}{\textbf{81.78}}} \\
			\multicolumn{2}{c|}{} & \multicolumn{2}{c|}{} & \multicolumn{2}{c}{} \\
			\hline
			\multicolumn{2}{c|}{\multirow{2}[1]{*}{Kernel Function}} & \multicolumn{2}{c|}{\multirow{2}[1]{*}{EAA mAP}} & \multicolumn{2}{c}{\multirow{2}[1]{*}{DC mAP}} \\
			\multicolumn{2}{c|}{} & \multicolumn{2}{c|}{} & \multicolumn{2}{c}{} \\
			\hline
			\multicolumn{2}{c|}{\multirow{2}[1]{*}{Epanechnikov}} & \multicolumn{2}{c|}{\multirow{2}[1]{*}{58.73}} & \multicolumn{2}{c}{\multirow{2}[1]{*}{80.68}} \\
			\multicolumn{2}{c|}{} & \multicolumn{2}{c|}{} & \multicolumn{2}{c}{} \\
			\multicolumn{2}{c|}{\multirow{2}[0]{*}{Uniform}} & \multicolumn{2}{c|}{\multirow{2}[0]{*}{56.28}} & \multicolumn{2}{c}{\multirow{2}[0]{*}{81.31}} \\
			\multicolumn{2}{c|}{} & \multicolumn{2}{c|}{} & \multicolumn{2}{c}{} \\
			\multicolumn{2}{c|}{\multirow{2}[0]{*}{Triangle}} & \multicolumn{2}{c|}{\multirow{2}[0]{*}{50.97}} & \multicolumn{2}{c}{\multirow{2}[0]{*}{72.17}} \\
			\multicolumn{2}{c|}{} & \multicolumn{2}{c|}{} & \multicolumn{2}{c}{} \\
			\multicolumn{2}{c|}{\multirow{2}[0]{*}{Cosine}} & \multicolumn{2}{c|}{\multirow{2}[0]{*}{56.36}} & \multicolumn{2}{c}{\multirow{2}[0]{*}{74.14}} \\
			\multicolumn{2}{c|}{} & \multicolumn{2}{c|}{} & \multicolumn{2}{c}{} \\
			\multicolumn{2}{c|}{\multirow{2}[0]{*}{Gauss}} & \multicolumn{2}{c|}{\multirow{2}[0]{*}{\textbf{60.02}}} & \multicolumn{2}{c}{\multirow{2}[0]{*}{\textbf{81.78}}} \\
			\multicolumn{2}{c|}{} & \multicolumn{2}{c|}{} & \multicolumn{2}{c}{} \\
			\hline
	\end{tabular}}
	
\end{table}%

\begin{table}[t]
	\centering
	\small
	\caption{Ablation study on the settings of RCM module. 	\label{als_fusion}}
	\renewcommand{\arraystretch}{0.7}
	\setlength{\tabcolsep}{3mm}{
		\begin{tabular}{cc|cc|cc}
			\hline
			\multicolumn{2}{c|}{\multirow{2}[1]{*}{Compensation Source}} & \multicolumn{2}{c|}{\multirow{2}[1]{*}{EAA mAP}} & \multicolumn{2}{c}{\multirow{2}[1]{*}{DC mAP}} \\
			\multicolumn{2}{c|}{} & \multicolumn{2}{c|}{} & \multicolumn{2}{c}{} \\
			\hline
			\multicolumn{2}{c|}{\multirow{2}[1]{*}{Radar and Vision}} & \multicolumn{2}{c|}{\multirow{2}[1]{*}{57.12}} & \multicolumn{2}{c}{\multirow{2}[1]{*}{77.88}} \\
			\multicolumn{2}{c|}{} & \multicolumn{2}{c|}{} & \multicolumn{2}{c}{} \\
			\multicolumn{2}{c|}{\multirow{2}[0]{*}{Vision Only}} & \multicolumn{2}{c|}{\multirow{2}[0]{*}{56.06}} & \multicolumn{2}{c}{\multirow{2}[0]{*}{77.32}} \\
			\multicolumn{2}{c|}{} & \multicolumn{2}{c|}{} & \multicolumn{2}{c}{} \\
			\multicolumn{2}{c|}{\multirow{2}[1]{*}{Radar Only}} & \multicolumn{2}{c|}{\multirow{2}[1]{*}{\textbf{60.02}}} & \multicolumn{2}{c}{\multirow{2}[1]{*}{\textbf{81.78}}} \\
			\multicolumn{2}{c|}{} & \multicolumn{2}{c|}{} & \multicolumn{2}{c}{} \\
			\hline
			\multicolumn{2}{c|}{\multirow{2}[1]{*}{Receptive Field Combination}} & \multicolumn{2}{c|}{\multirow{2}[1]{*}{EAA mAP}} & \multicolumn{2}{c}{\multirow{2}[1]{*}{DC mAP}} \\
			\multicolumn{2}{c|}{} & \multicolumn{2}{c|}{} & \multicolumn{2}{c}{} \\
			\hline
			\multicolumn{2}{c|}{\multirow{2}[1]{*}{[3, 5, 7]}} & \multicolumn{2}{c|}{\multirow{2}[1]{*}{59.93}} & \multicolumn{2}{c}{\multirow{2}[1]{*}{81.53}} \\
			\multicolumn{2}{c|}{} & \multicolumn{2}{c|}{} & \multicolumn{2}{c}{} \\
			\multicolumn{2}{c|}{\multirow{2}[0]{*}{[5, 7, 9]}} & \multicolumn{2}{c|}{\multirow{2}[0]{*}{59.34}} & \multicolumn{2}{c}{\multirow{2}[0]{*}{80.87}} \\
			\multicolumn{2}{c|}{} & \multicolumn{2}{c|}{} & \multicolumn{2}{c}{} \\
			\multicolumn{2}{c|}{\multirow{2}[0]{*}{[3, 7, 9]}} & \multicolumn{2}{c|}{\multirow{2}[0]{*}{58.22}} & \multicolumn{2}{c}{\multirow{2}[0]{*}{79.74}} \\
			\multicolumn{2}{c|}{} & \multicolumn{2}{c|}{} & \multicolumn{2}{c}{} \\
			\multicolumn{2}{c|}{\multirow{2}[0]{*}{[3, 5, 9]}} & \multicolumn{2}{c|}{\multirow{2}[0]{*}{\textbf{60.02}}} & \multicolumn{2}{c}{\multirow{2}[0]{*}{\textbf{81.78}}} \\
			\multicolumn{2}{c|}{} & \multicolumn{2}{c|}{} & \multicolumn{2}{c}{} \\
			\hline
	\end{tabular}}
	
\end{table}

\begin{table}[t]
	\centering
	\small
	\caption{Ablation study on the settings of MMIF module. 	\label{als_dmf}}
	\renewcommand{\arraystretch}{0.7}
	\setlength{\tabcolsep}{3mm}{
		\begin{threeparttable}	
			\begin{tabular}{cc|cc|cc}
				\hline
				\multicolumn{2}{c|}{\multirow{2}[1]{*}{Channel Transformation}} & \multicolumn{2}{c|}{\multirow{2}[1]{*}{EAA mAP}} & \multicolumn{2}{c}{\multirow{2}[1]{*}{DC mAP}} \\
				\multicolumn{2}{c|}{} & \multicolumn{2}{c|}{} & \multicolumn{2}{c}{} \\
				\hline
				\multicolumn{2}{c|}{\multirow{2}[1]{*}{MeanPooling + Linear}} & \multicolumn{2}{c|}{\multirow{2}[1]{*}{57.42}} & \multicolumn{2}{c}{\multirow{2}[1]{*}{78.69}} \\
				\multicolumn{2}{c|}{} & \multicolumn{2}{c|}{} & \multicolumn{2}{c}{} \\
				\multicolumn{2}{c|}{\multirow{2}[0]{*}{MaxPooling + Linear}} & \multicolumn{2}{c|}{\multirow{2}[0]{*}{56.83}} & \multicolumn{2}{c}{\multirow{2}[0]{*}{78.51}} \\
				\multicolumn{2}{c|}{} & \multicolumn{2}{c|}{} & \multicolumn{2}{c}{} \\
				\multicolumn{2}{c|}{\multirow{2}[0]{*}{CBR*}} & \multicolumn{2}{c|}{\multirow{2}[0]{*}{57.70}} & \multicolumn{2}{c}{\multirow{2}[0]{*}{77.55}} \\
				\multicolumn{2}{c|}{} & \multicolumn{2}{c|}{} & \multicolumn{2}{c}{} \\
				\multicolumn{2}{c|}{\multirow{2}[0]{*}{Reshape}} & \multicolumn{2}{c|}{\multirow{2}[0]{*}{\textbf{60.02}}} & \multicolumn{2}{c}{\multirow{2}[0]{*}{\textbf{81.78}}} \\
				\multicolumn{2}{c|}{} & \multicolumn{2}{c|}{} & \multicolumn{2}{c}{} \\
				\hline
				\multicolumn{2}{c|}{\multirow{2}[1]{*}{Mamba Type}} & \multicolumn{2}{c|}{\multirow{2}[1]{*}{EAA mAP}} & \multicolumn{2}{c}{\multirow{2}[1]{*}{DC mAP}} \\
				\multicolumn{2}{c|}{} & \multicolumn{2}{c|}{} & \multicolumn{2}{c}{} \\
				\hline
				\multicolumn{2}{c|}{\multirow{2}[1]{*}{None}} & \multicolumn{2}{c|}{\multirow{2}[1]{*}{57.68}} & \multicolumn{2}{c}{\multirow{2}[1]{*}{79.31}} \\
				\multicolumn{2}{c|}{} & \multicolumn{2}{c|}{} & \multicolumn{2}{c}{} \\
				\multicolumn{2}{c|}{\multirow{2}[0]{*}{Vanilla}} & \multicolumn{2}{c|}{\multirow{2}[0]{*}{58.14}} & \multicolumn{2}{c}{\multirow{2}[0]{*}{80.82}} \\
				\multicolumn{2}{c|}{} & \multicolumn{2}{c|}{} & \multicolumn{2}{c}{} \\
				\multicolumn{2}{c|}{\multirow{2}[0]{*}{Max-Mamba}} & \multicolumn{2}{c|}{\multirow{2}[0]{*}{59.97}} & \multicolumn{2}{c}{\multirow{2}[0]{*}{81.43}} \\
				\multicolumn{2}{c|}{} & \multicolumn{2}{c|}{} & \multicolumn{2}{c}{} \\
				\multicolumn{2}{c|}{\multirow{2}[0]{*}{Mean-Mamba}} & \multicolumn{2}{c|}{\multirow{2}[0]{*}{\textbf{60.02}}} & \multicolumn{2}{c}{\multirow{2}[0]{*}{\textbf{81.78}}} \\
				\multicolumn{2}{c|}{} & \multicolumn{2}{c|}{} & \multicolumn{2}{c}{} \\
				\hline
			\end{tabular}
			\begin{tablenotes}
				\item { $*$ represents Convolution + BatchNorm + ReLU.}
			\end{tablenotes} 
		\end{threeparttable}
		
	}
	
\end{table}%

\begin{figure*}[t]
	\centering
	\includegraphics[width=6.5in]{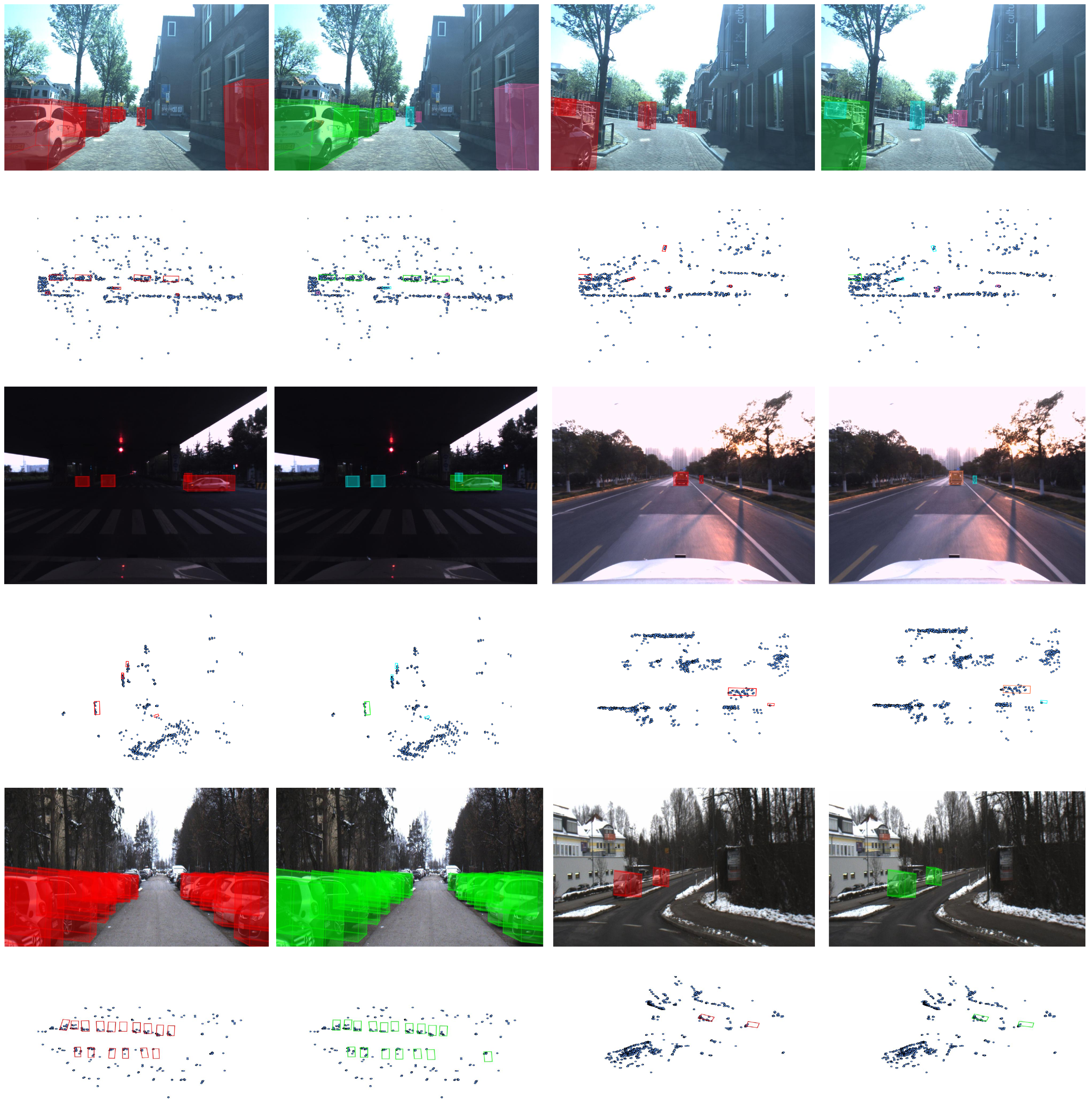}
	\caption{The visualization results of SDCM. There are two scenes for every dataset, with two rows which are the visualizations on RGB images and under point cloud BEV perspective. The red, green, blue, pink and orange 3-D bounding boxes denote GroundTruth, Car, Cyclist, Pedestrian and Truck respectively. The scenes contains low-light, long distance detection and dense occlusion.\label{det}
	}
\end{figure*}
Fig \ref{surface} (c) shows the quantity statistics of points in BEV pillars, corresponding to three scenes in Fig \ref{pc1}.
Notably, The size of pillars is 1 m $\times$ 1 m.
 The left sub-figures denote the original point clouds, while the right sub-figures denote the point clouds after densifying. It could be seen that the under BEV perspective, the objects contains more measurment information, which could be conveyed to the pseudo images generated by pillar encoding mechanism.

These experiment results shows that our densifying could enhance the differenr type of feature learning input like 3-D point clouds, BEV images and 2-D projected point clouds.

\subsubsection{Ablation Study of RCM module }
The ablation study results of the settings of RCM module are shown in Table \ref{als_fusion}. The objects of this ablation study are compensation source and receptive field combination. Notably, for different compensation source, similar to RCFusion \cite{RCFusion}, the features of Image Only and Radar Only are concatenated and conveyed to CBR (Convolution-BatchNorm-ReLU) module for fusion.
It could be seen that only adopting radar features ad compensation source achieves the best performance. However, when adopting vision features as compensation source, the metrics decrease about 3.96 and 4.46 points, mainly because image features could not provide more information for radar features under poor lighting conditions. When adopting both of two features as compensation sources, the metrics have minor improvement but they are lower about 2.90 and 3.90 points than Radar only.
This experiment result shows that it is important to adopt the radar features to achieve compensation of representation degradation during fusion. 

The affects of receptive field combination are shown in Table \ref{als_fusion}. From the result it could be seen that the detection results of [3, 5, 7], [5, 7, 9]  and [3, 5, 9] 
are approximate. Only [3, 7, 9] has obvious biases, comapring with the results of [3, 5, 9]. The reason may be that [3, 7, 9] has relatively wide average receptive field range. Following these results, [3, 5, 9] is chosen for RCM module.

\subsubsection{Ablation Study of MMIF module }
Table \ref{als_dmf} shows the ablation study results of MMIF module. The channel transformation and the type of Mamba block are the objects of ablation study.
For channel transformation method, it could be seen that Reshape could achieve the best performance. Other three replacements have to adopt unpooling or transposed convolution to recover the channel after Mamba blocks, which may damage to complementarity, such as local texture features. For the type of Mamba block, it could be seen that the Max-Mamba and Mean-Mamba achieve similar effects. Comparing with the vanilla version, the Mean-Mamba has about 1.88 and 0.96 points improvement in EAA and DC mAP respectively, because it obtains deep global features under the assistance of channel information. The performance is poor without Mamba block because the channel transformation reduces the resolution of feature maps, whose long range dependecies need to be captured by Mamba mechanism again.

\subsection{Visualization Results Analysis}
Fig \ref{det} shows the visualization results of proposed method on three datasets.
Notably, there are two scenes in every datasets with two rows, the first row is the visualization results on the RGB images, while the second rows is the visualization results under the point clouds BEV perspective. Every scene shows with two columns, the first column is the Groundtruth, while the second column is the detection result. The red, green, blue, pink and orange 3-D bounding boxes denote Groundtruth, Car, Cyclist, Pedestrian and Truck. The radar point clouds under BEV perspective is from original official radar datas.

The scenes contains low-light, long distance detection and dense occlusion. The first scene of TJ4DRadSet shows that SDCM could capture the objects in low-light scenes. In addition, for long distance detection, such as all the scenes of VoD and the second scene of TJ4DRadSet dataset. The pedestrian, cyclist and truck are detected although they only occupied less pixels on the RGB images.  Our method also achieve dense occlusion detection, as shown in the first scenes of VoD and Astyx HiRes 2019 dataset. In the common scene, our method could detect the object correctly, such as the second scene of Astyx HiRes 2019 dataset.
These results are related to SimDen module which provide more point cloud information about them, as well as RCM and MMIF module, which achieve the compensation of representation degradation and the reduction of heterogeneous features.

\section{Conclusion}
In this article, we propose SDCM, which contains simulated densifying and compensatory modeling fusion for 3-D object detection in IoV. First, SimDen module is designed to achieve 4-D radar point clouds densifying, which faces the challege about the sparsity of radar point clouds. Second,
RCM module is designed to reduce the affect of representation degradation from vision datas. Third,  MMIF module is designed to reduce the heterogenity and achieve feature interaction fusion. The experiment results show that SDCM achieves outstanding performance on the precision of 3-D object detection, with less parameter quantity and faster inference speed.

\bibliographystyle{IEEEtran}
\bibliography{New_IEEEtran_how-to}

\end{document}